\documentclass{article} %
\usepackage{iclr2026_conference,times}

\usepackage{amsmath,amsfonts,bm}

\def\eqref#1{equation~\ref{#1}}

\def\1{\bm{1}}

\DeclareMathAlphabet{\mathsfit}{\encodingdefault}{\sfdefault}{m}{sl}
\SetMathAlphabet{\mathsfit}{bold}{\encodingdefault}{\sfdefault}{bx}{n}

\usepackage{hyperref}
\usepackage{url}
\usepackage{graphicx}
\usepackage{booktabs}
\usepackage[T1]{fontenc}
\usepackage{inconsolata}     
\usepackage{listings}
\usepackage{caption}
\usepackage{algorithm}
\usepackage{algpseudocode}
\usepackage{amsmath}
\usepackage{placeins}
\usepackage{wrapfig}

\algrenewcommand\algorithmicrequire{\textbf{Input:}}
\algrenewcommand\algorithmicensure{\textbf{Output:}}

\lstdefinestyle{prompt}{
  basicstyle=\ttfamily\small,
  frame=single,
  breaklines=true,
  columns=fullflexible,
  keepspaces=true,
  numbers=none,
  xleftmargin=0pt, xrightmargin=0pt
}

\usepackage{xcolor}
\usepackage{tcolorbox}
\usepackage{enumitem} 
\tcbuselibrary{breakable} 
\definecolor{promptgray}{gray}{0.97}

\newtcolorbox{systemprompt}[2][]{
  breakable,
  colback=promptgray,
  colframe=black,
  fonttitle=\bfseries,
  title=#2,
  #1
}

\lstdefinestyle{jsonstyle}{
  basicstyle=\ttfamily\small,  %
  breaklines=true,            %
  breakatwhitespace=true,     %
  showstringspaces=false,     %
  stringstyle=\color{jsonstring},
  keywordstyle=\color{jsonkey},
  otherkeywords={true,false,null}, %
  keywordstyle=[2]\color{jsonbool},
  numberstyle=\color{jsonnumber},
}

\title{\fullname: Scaling Up Memory for Robot\\ Control via Experience Retrieval}

\author{
\hspace{-0.25em}Ajay Sridhar\thanks{Equal contribution} \quad Jennifer Pan$^{*}$ \quad Satvik Sharma \quad Chelsea Finn \\
\normalfont Stanford University \\
\texttt{\{ajaysri, jrpan\}@stanford.edu}
}

\newcommand{\fullname}{MemER}
\newcommand{\website}{\url{https://jen-pan.github.io/memer/}}
\iclrfinalcopy %

\begin{document}

\maketitle

\begin{abstract}
Humans routinely rely on memory to perform tasks, yet most robot policies lack this capability; our goal is to endow robot policies with the same ability. Naively conditioning on long observation histories is computationally expensive and brittle under covariate shift, while indiscriminate subsampling of history leads to irrelevant or redundant information. We propose a hierarchical policy framework, where the high-level policy is trained to select and track previous relevant keyframes from its experience. The high-level policy uses selected keyframes and the most recent frames when generating text instructions for a low-level policy to execute. This design is compatible with existing vision-language-action (VLA) models and enables the system to efficiently reason over long-horizon dependencies. 
In our experiments, we finetune Qwen2.5-VL-7B-Instruct and $\pi_{0.5}$ as the high-level and low-level policies respectively, using demonstrations supplemented with minimal language annotations. Our approach, \fullname, outperforms prior methods on three real-world long-horizon robotic manipulation tasks that require minutes of memory. Videos and code can be found at \website.
\end{abstract}

\section{Introduction}

In recent times, we have seen significant strides in the language-following and generalization capabilities of robotic manipulation policies \citep{rt2, intelligence2025pi05visionlanguageactionmodelopenworld, openvla, gr00t1, geminiroboticsteam2025geminirobotics15pushing}. While these policies are improving for real-world deployment, a critical limitation remains: the absence of long-term memory. Memory allows humans to handle the inherent partial observability found in their environment. For instance, if a person wanted to make a sandwich, they would have to recall where they saw the jar of peanut butter or the knife, especially if these items had not been recently viewed. The ability to form and retrieve memories is a crucial step towards robots solving complex, multi-step tasks. The goal of this paper is to provide an effective way to enable existing generalist policies to solve tasks that require long-term visual memory. \looseness=-1%

Because conditioning on long sequences of high-dimensional image and video sequences is computationally expensive, many existing generalist end-to-end policies are trained with little to no visual history. The high memory cost makes training prohibitively expensive and model deployment unusably slow. Furthermore, long observation histories can often introduce a form of overfitting — shortcut reliance on spurious correlations between inputs and demonstrator actions \citep{ptp}. The policy misgeneralizes under its own state distribution, leading to performance degradation during deployment due to compounding covariate shift between states visited by the demonstrator policy and the learned policy. The suboptimal policy will generate histories that differ from those seen by the expert, which is only made worse as the observation history lengthens.

Some past works have shown it is possible to expand the observation context of their policy via auxiliary losses \citep{ptp}, or by finetuning pretrained foundation models for action prediction with native memory capabilities \citep{sam2act}. Although these methods significantly increase the types of tasks a robot can execute, they are challenging to naively scale to long histories. %
To overcome this, policies must learn to filter out and store task-relevant information from the full historical context to prevent the memory footprint from exploding on tasks that require long-range dependencies.

To this end, we propose approaching long-term memory for robotic policies with a hierarchical framework. The high-level policy is a finetuned video-understanding VLM trained to output subtasks and, most importantly, to select keyframes from its fixed recent context that represent important information it will need to remember to solve the task. The low-level policy is a generalist robot policy finetuned to execute the subtask specified by the high-level policy, handling the robot-specific challenges of the task that require high-frequency inference such as kinematic control. We take advantage of the fact that these open-source VLMs are finetuned on large amounts of video understanding data. With this strong prior, we find that we only need 50 teleoperated robot demonstrations with subtask annotations to adapt these VLMs to accomplish robot-specific memory-based tasks \citep{qwen2.5}.

Our contribution is \fullname, a framework for scaling up \textbf{Mem}ory in robotic control  via \textbf{E}xperience \textbf{R}etrieval. We demonstrate \fullname's ability to effectively utilize task-relevant past information on three complex long-horizon tasks that require up to a few of minutes of memory. To the best of our knowledge, our real-world manipulation tasks necessitate reasoning over longer horizons than prior work. \looseness=-1

\begin{figure}[t!]
  \centering
  \includegraphics[width=\linewidth]{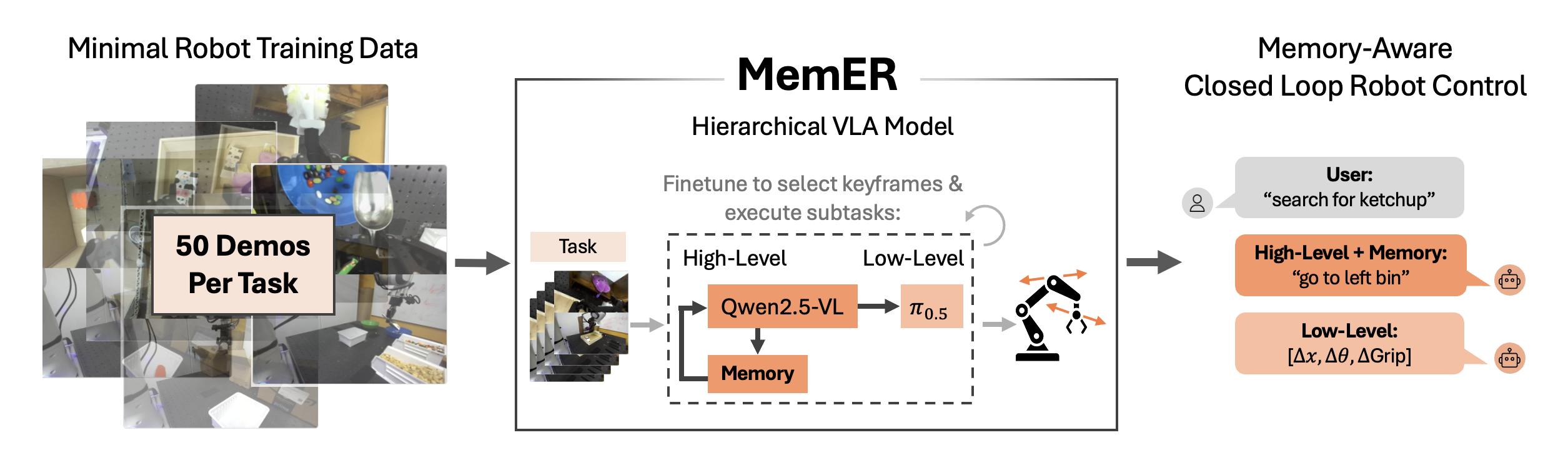}
  \caption{\textbf{MemER Overview.} We present MemER, a framework for scaling up memory in robotic control. MemER is able to utilize task-relevant past information to effectively reason through three complex long-horizon tasks, using a single policy trained with only a small number of expert demonstrations.}
  \label{fig:fig1}
\end{figure}

\section{Related Work}
\textbf{Memory for Long-Context Robot Policies.}
Memory is essential for generalist robots to complete complex tasks. Prior work primarily studies memory in the context of comparatively short-horizon tasks. For example, \cite{ptp} and \cite{sam2act} use different approaches to extend the context of imitation policies from a few frames to at most two dozen. We further investigate tasks which require building memory from hundreds of frames. Similar to our approach, \cite{sam2act} stores previous frames in memory. However, these are at most the $N=10$ most recent frames of context, while our method can choose to include task-relevant frames spanning the entire episode, which could contain over a thousand frames. 
Another body of work investigates the compression of images in the policy's context. \cite{litevlp} compresses similar observations in pixel space, which works well for stationary cameras but not for wrist-mounted cameras, which are necessary for many dexterous robotic tasks. \cite{zheng2025tracevlavisualtraceprompting} compresses all the past observations into 2D visual traces of the motion of the end-effector and moving objects in the scene. Traces help the model reason about recent spatiotemporal changes, but this approach is hard to extend tasks that require long-term memory. Memory has also played a major role in robotic navigation research. Some navigation works represent memory with an explicit geometric and/or semantic map of the environment \cite{ijrr2012henry, legs}. Spatial maps of the environment are hard to apply in manipulation tasks since the robot often modifies the environment. Other works directly prompt API-Based VLMs with video context to decide where the robot should navigate \citep{mobilityvla, sms, vlmpc}. We found that existing API-Based VLMs are not sufficient to reason about robot affordances for our long-horizon manipulation tasks, so we resort to finetuning open-weight models.

\textbf{Foundation Models in Robotics.}
Recent progress in vision-language-action models (VLAs) have allowed for impressive generality in robotics. VLAs combine web-scale pretraining with expressive action decoding mechanisms to execute real-world tasks. Most VLAs usually follow one of two paradigms: 1) single end-to-end model which takes images and a language task as input and outputs actions \citep{black2024pi0visionlanguageactionflowmodel, rt2} or 2) a high-level policy that takes images and a language task as input and outputs some intermediate embedding, language subtask or waypoints, which a lower-level policy conditions on when outputting actions \citep{yayrobot, hamster, shentu2024lcb}. 
Our work builds on the second paradigm; we train the high-level policy to incorporate a keyframe filter so it can predict accurate language subtasks for the low-level policy to execute. %

\textbf{Video Keyframe Selection.}
Outside of robotics, previous work in computer vision has also studied incorporating longer contexts for VLMs \citep{amego, ESOM}. Similar to our work, other works have used keyframe selection to improve video understanding and question answering \citep{sevila, mvu}. Many such methods incur a high per-frame cost because they estimate frame importance via separate multimodal-LLM calls. These methods are not directly applicable for robotic tasks because increasing video context lengths would not meet the task's latency constraint during inference. \cite{hu2025mllmbasedvideoframe} uses lightweight models to score all frames in a single pass, which reduces per-frame cost but lacks the ability to continuously stream image observations. Departing from existing VLM work for VQA, \cite{hu2025mllmbasedvideoframe} uses non-uniform frame sampling through a lightweight scoring model; in contrast, we achieve non-uniform sampling without additional models. Designed for real-world robotics, our method emphasizes low-cost inference and streaming support. %

\section{\fullname}

\subsection{Preliminaries}
\label{sec:prelim}
\textbf{Language-Conditioned Control Policies.} Language-conditioned robot policies are typically trained to model the conditional distribution $\pi(\boldsymbol{A_t}|o_t)$, where $\boldsymbol{A_t} = [a_t, a_{t+1}, \ldots a_{t+H-1}]$ is a chunk of actions modeled from the current timestep $t$, up to $H$ timesteps in the future 
\citep{zhao2023learningfinegrainedbimanualmanipulation} and $o_t$ is the robot's current sensor observation. The current observation is usually formulated as $o_t = [\boldsymbol{I_t}, \mathit{l_t}, \boldsymbol{q_t}]$, where $\boldsymbol{I_t} = [I^1_t, I^2_t, \ldots, I^n_t]$ are images from multiple cameras, $\mathit{l_t}$ is the language instruction, and $\boldsymbol{q_t}$ are the proprioceptive inputs from the robot (i.e. joint angles and gripper state) \citep{black2024pi0visionlanguageactionflowmodel, octomodelteam2024octoopensourcegeneralistrobot}. 

\textbf{Memory-Based Tasks.} We consider a set of tasks for which the robot policy must leverage past information to successfully complete due to partial observability in the environment. In other words, a robot policy trained to model $\pi(\boldsymbol{A_t}|o_t)$
could not complete the task, but a policy trained to model $\pi(\boldsymbol{A_t}|o_{0:t})$ could.

\textbf{Hierarchical Policies.} In order to execute complex, long-horizon tasks, we follow \cite{hirobot} and hierarchically decompose the robot policy into a low-level control policy ($\pi_l$) and a high-level policy ($\pi_h$) that generates instructions. 
\begin{equation}
    \pi(\boldsymbol{A_t}|o_t) = \pi_l(\boldsymbol{A_t}|[\boldsymbol{I_t}, \mathit{l_t'}, \boldsymbol{q_t}])\pi_h(l_t'|\boldsymbol{I_t}, \mathit{l_t)}
\end{equation}

The high-level policy models the conditional distribution $\pi_h(l_t'|\boldsymbol{I_t}, l_t)$, where $l_t'$ is the current language subtask that the low-level policy conditions on to complete the overall instruction $l_t$. The high-level policy could be represented as a separate VLM \citep{hirobot} or share the same weights as the low-level control policy \citep{intelligence2025pi05visionlanguageactionmodelopenworld}. In our method, the high-level policy will be responsible for reasoning about memory. %

\textbf{Data Collection for Hierarchical Policies.}
Similar to prior work, we use language subtasks $\mathit{l_t'}$ to label each observation in a trajectory \citep{yayrobot, hirobot, intelligence2025pi05visionlanguageactionmodelopenworld}. We end up with a dataset of trajectories of the following tuple $(\boldsymbol{I_t}, \boldsymbol{q_t}, \mathit{l_t}, \mathit{l_t'},a_t)$ to train our high-level and low-level policies. For example, we can take the task "search for ketchup" and break it down into the following subtasks: "look in left bin", "look in right bin", and "take out ketchup from right bin." Examples of the subtasks and image observations for each task are shown in Figure \ref{fig:tasksubtasks}. 

In our data-collection setup, the operator executes a prescribed subtask and presses a key upon completion to advance to the next subtask. Consistent with previous work, we supplement the low-level policy training set with 10–15 intervention demonstrations to improve robustness at deployment \citep{rac, kim2025srthhierarchicalframeworkautonomous}. Since the low-level policy is Markovian, we can efficiently collect the intervention data by initializing the robot in common failure states we have observed during deployment. Then, we teleoperate the robot back into an in-distribution state.

\definecolor{green}{RGB}{1, 168, 157}
\definecolor{pink}{RGB}{203, 41, 123}
\begin{figure}[t!]    
    \centering
    \includegraphics[width=0.8\linewidth]{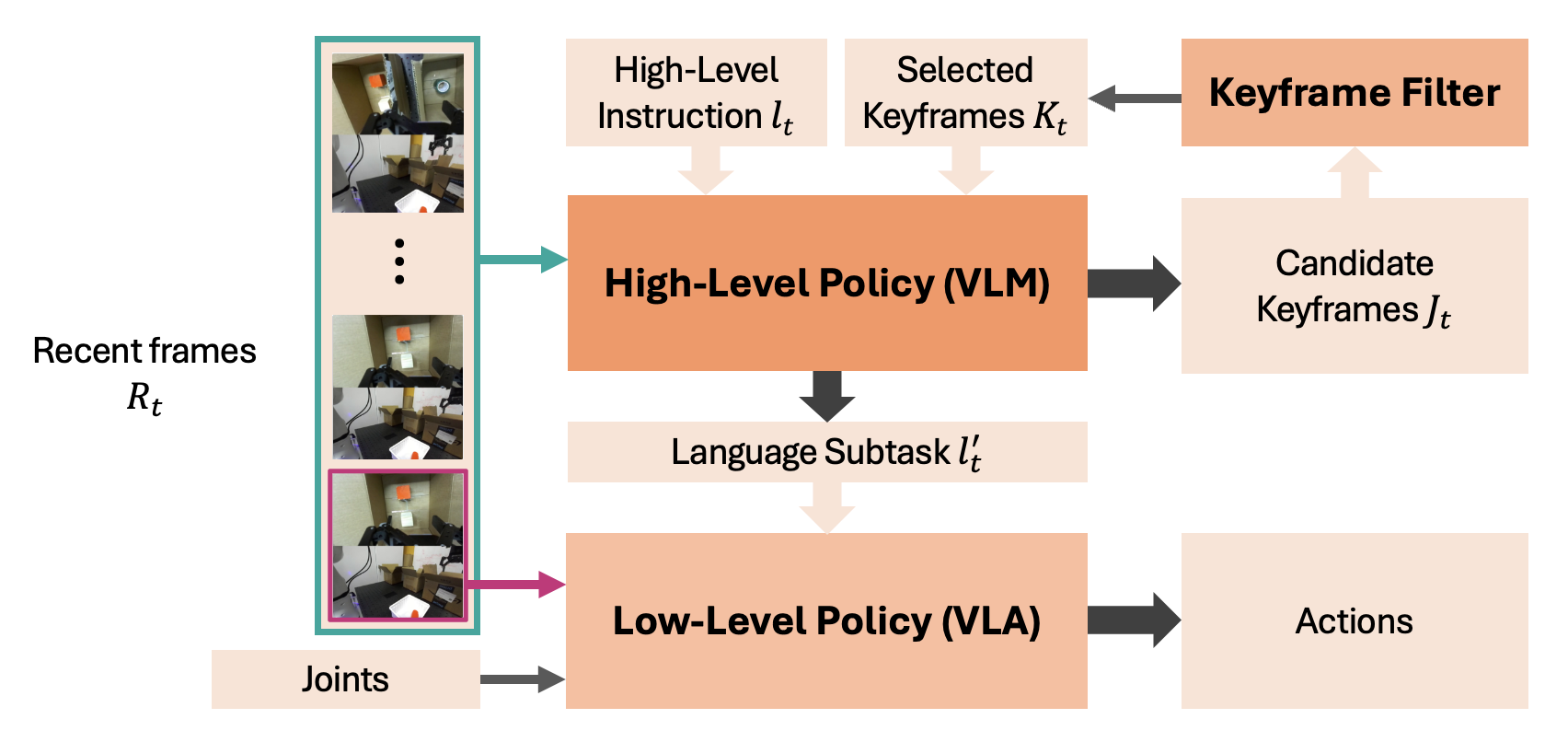}
    \vspace{-10pt}
    \caption{\textbf{Architecture of \fullname.} The high-level policy processes task instructions, selected keyframes (if any), and \textcolor{green}{recent images} from base and wrist-mounted cameras to generate low-level language subtasks and candidate keyframes (if any). The low-level policy uses the subtask, \textcolor{pink}{current image}, and robot joint states to produce actions. The candidate keyframe(s) are processed by the keyframe filter to obtain the selected keyframes for input during the next step of inference.}
    \label{fig:architecture}
\end{figure}

\subsection{High-Level Policy}
\label{sec:high-level-policy}
Our method builds on the common hierarchical VLA paradigm, as in \citet{hirobot}, and extends it with the ability to tackle long-horizon tasks that require memory. We choose to adopt a hierarchical apporach because open source VLMs like Qwen2.5-VL-7B-Instruct have a strong video understanding prior from the video datasets they have been trained on, and thus can be adapted for memory-based planning \citep{qwen2.5}. We use a finetuned VLM as the high-level policy to both nominate candidate keyframes and predict subtasks for the low-level policy during closed-loop control, as shown in Figure \ref{fig:architecture}.
The candidate keyframes are then filtered for redundancy and added to a group of selected keyframes that the high-level policy  conditions on continuously when predicting the next subtask and candidate keyframes.
Concretely, at each timestep, we feed our high-level memory policy 1) the last $N$ frames per camera $\boldsymbol{R_t} = \boldsymbol{I_{t-N+1:t}}$, where $N$ is the integer context-window shared across cameras, 
2) the high-level task instruction $\mathit{l_t}$, and 3) previously selected keyframes $\boldsymbol{K_t} \subseteq \boldsymbol{I_{0:t-N+1}}$, where practically $|K_t| \leq 8$. 
The high-level policy then predicts two things: (i) the current subtask to execute $\mathit{l_t'}$ and (ii) 
 the candidate keyframes $\boldsymbol{J_t} \subseteq \boldsymbol{R_t}$, a subset of frames from the recent context.
All together, our high-level policy models $\pi_h(l_t', \boldsymbol{J_t} | \boldsymbol{R_t}, \boldsymbol{K_t})$. 
The low-level policy conditions on $l'_t$ to predict the direct joint velocities for the robot as described in Section \ref{sec:pratical_implementation}.
In parallel, $\boldsymbol{K_{t+1}}$, the selected keyframes for the next timestep of high-level inference, are calculated from the sequence of all candidate keyframes predicted since the start of the trajectory $\boldsymbol{J'_{0:t}} = (\boldsymbol{J_0},\boldsymbol{J_1}, \ldots, \boldsymbol{J_t})$ using a simple 1D single-linkage clustering algorithm described in the following paragraphs. 

\begin{figure}[t]
\centering
\includegraphics[width=\linewidth]{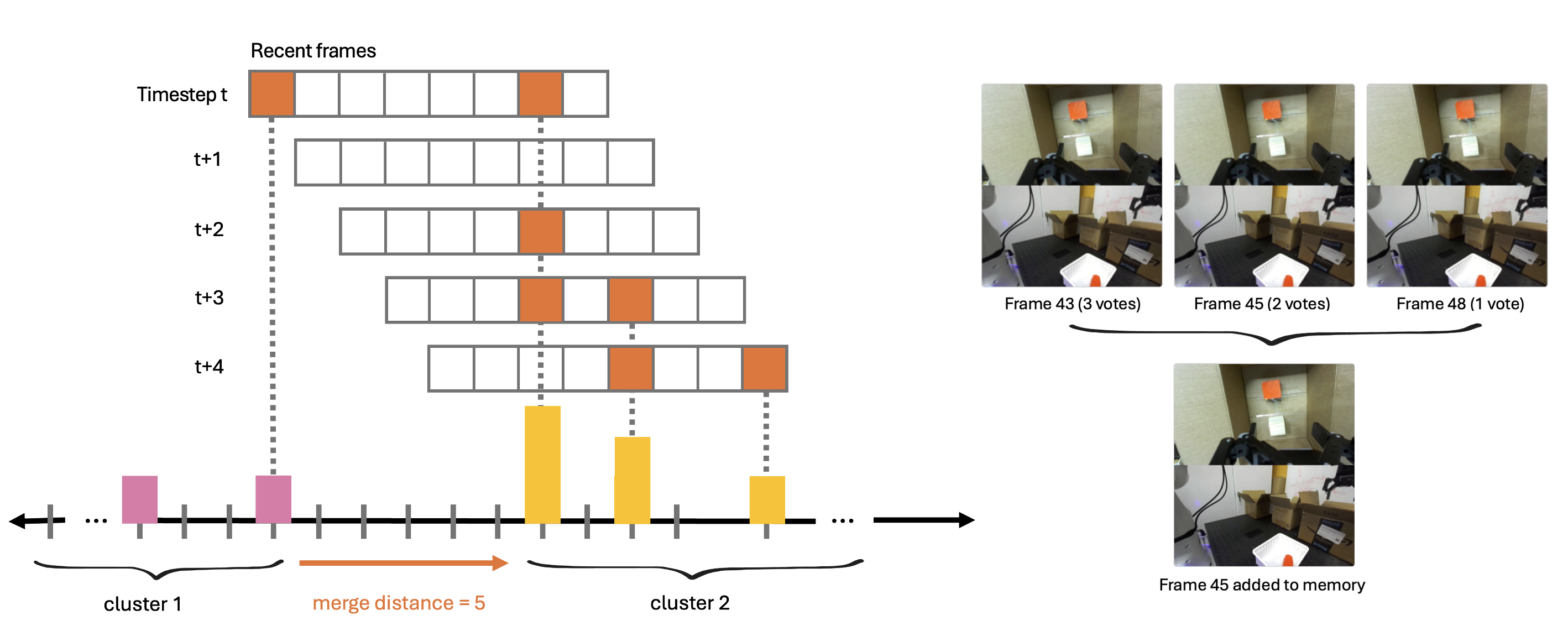} 
\caption{\textbf{1D single-linkage over nominated frames.} At each timestep, the high-level policy nominates candidate keyframe(s), as highlighted in orange. All candidate keyframes are aggregated across time with 1D single-linkage using a merge distance of $d=5$ frames, yielding disjoint clusters. For each cluster, the colored bars indicate nominations for the observation at that timestamp, with bar height proportional to the number of nominations received. We select one representative frame per cluster by taking the median keyframe of all the candidates, and add that frame to memory.}
\label{fig:clustering}
\end{figure}

\textbf{Building Visual Memory.} To build visual memory, our keyframe filter consolidates observations that have just exited our recent context, $\boldsymbol{R_t}$, into our selected keyframes $\boldsymbol{K_t}$. The filtering process operates on the temporal indices of these keyframes. The approach is task-agnostic and allows us to have coverage of all frames in the stream. By timestep $t$, the high-level policy has given us $\boldsymbol{J'_{0:t}} = (\boldsymbol{J_i})_{i=0}^t$, the sequence of candidate sets nominated up to timestep $t$. We begin by extracting the temporal index of every frame nominated in this sequence and pooling them into a single temporally ordered list $\boldsymbol{G_{0:t}}$. Importantly, we preserve duplicate indices, as this allows the subsequent median selection to aggregate repeated nominations and yield the most representative frame from each cluster.

We then create clusters, $\boldsymbol{C_i}$, for all candidate indices in $\boldsymbol{G_{0:t}}$ by grouping keyframe indices that are at most $d$ apart from each other. For example, if $\boldsymbol{G_{0:t}}$ has temporal indices $\{1, 3, 3, 4, 10\}$ and $d=5$, then we would have two clusters $\boldsymbol{C_1} = \{1, 3, 3, 4\}$ and $\boldsymbol{C_2} =\{10\}$. After constructing the sequence of clusters, $(\boldsymbol{C_1}, \boldsymbol{C_2}, \ldots)$, we select the median index from each $\boldsymbol{C_i}$ to be that cluster's representative keyframe. The  keyframes corresponding to these final median indices represent $\boldsymbol{K_t}$, the selected keyframes for timestep $t$. For efficiency, clusters that have indices less than $t-N+1-d$ do not need to be recalculated. Figure \ref{fig:clustering} is a visualization of the clustering and keyframe selection during deployment. We provide detailed pseudocode in Appendix \ref{app:algo} and a video demonstration of the clustering algorithm in action on our project website \space\website.

\subsection{Practical Implementation of \fullname}
\label{sec:pratical_implementation}
\textbf{Training the Low-Level Policy.} For our low-level robot policy, we finetune a version of $\pi_{0.5}$ \citep{intelligence2025pi05visionlanguageactionmodelopenworld} trained on the DROID dataset \citep{khazatsky2025droidlargescaleinthewildrobot}. Given trajectories of $(\boldsymbol{I_t}, \boldsymbol{q_t}, \mathit{l_t'},a_t)$ we train our low-level policy to model the conditional distribution  $\pi_l(\boldsymbol{A_t} | \boldsymbol{I_t}, \boldsymbol{q_t}, \mathit{l_t'})$. We finetune the $\pi_{0.5}$ checkpoint trained on the DROID dataset due to its strong out-of-the-box behavior on the DROID setup we use for all of our experiments. Consequently, we find that we need only 50 demos of long-horizon trajectories and 10-15 examples of interventions for each of the three tasks to finetune a strong low-level policy. We finetune a single low-level policy on all three tasks. Refer to Appendix \ref{sec:finetuning_hyperparams} for the specific training parameters.

 \textbf{Training the High-Level Policy.} For our high-level policy, we finetune Qwen2.5-VL-7B-Instruct to predict two things: 1) the current subtask to execute and 2) any task-relevant keyframes to remember from the most recent frames (as described in section \ref{sec:high-level-policy}). We finetune a single high-level policy on all three tasks, as it gives the added benefit of stronger object generalization (see Appendix \ref{sec:cross-task-objs} for comparisons with the single-task variant of \fullname). We freeze the weights of the vision encoder and projection layer during finetuning for training efficiency. Refer to Appendix \ref{sec:finetuning_hyperparams} for the specific training parameters and Appendix \ref{app:prompts_for_training} for format of the training prompts.

\textbf{Annotating Keyframes for the High-Level Policy.} To label keyframes for each task, we employ a semi-automatic annotation procedure. First, we identify the transition points between consecutive subtasks and extract the boundary frames as candidate keyframes. Next, a human annotator reviews a small number of demonstrations to determine a simple annotation rule per subtask—specifying whether to select the \textbf{first}, \textbf{last}, or \textbf{no} frame within that subtask segment, depending on which option most effectively captures a visually informative state. For example, the rule may indicate selecting the last frame in "look inside the center bin", the first frame in "dust bottom shelf", or no frame for "reset scooper position." Once established, this rule is fixed per subtask type and automatically applied to all demonstrations of the corresponding task, yielding at most one keyframe per subtask segment. The resulting set of keyframes forms the ground-truth targets used to train the high-level policy. See Appendix \ref{sec:appendix_keyframe_rules} for the specific keyframe annotation rules for all of the subtasks in our three tasks.

\begin{figure}[t]
\centering

\begin{minipage}[t]{0.53\linewidth}
  \centering
  \includegraphics[width=\linewidth]{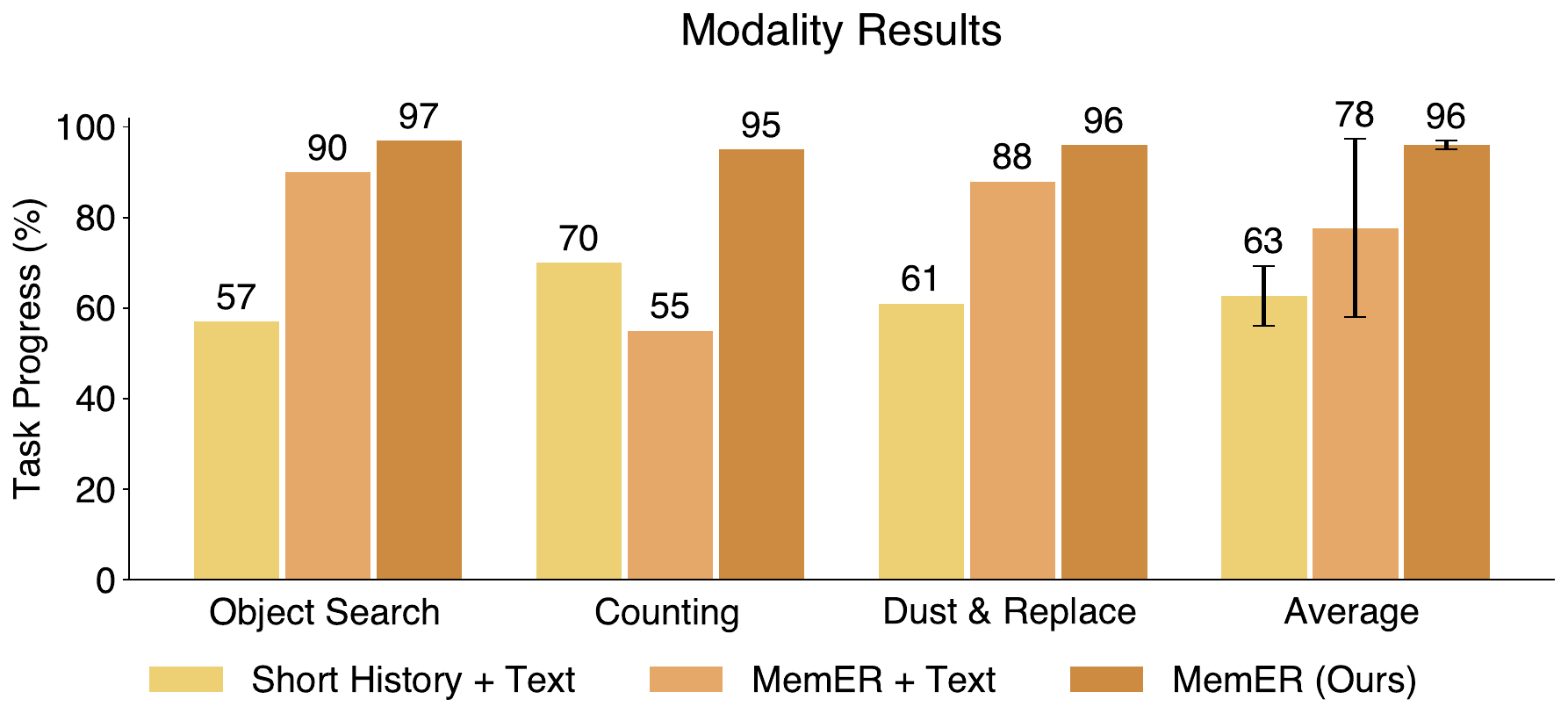}
\end{minipage}
\hfill %
\begin{minipage}[t]{0.45\linewidth}
  \centering
  \includegraphics[width=\linewidth]{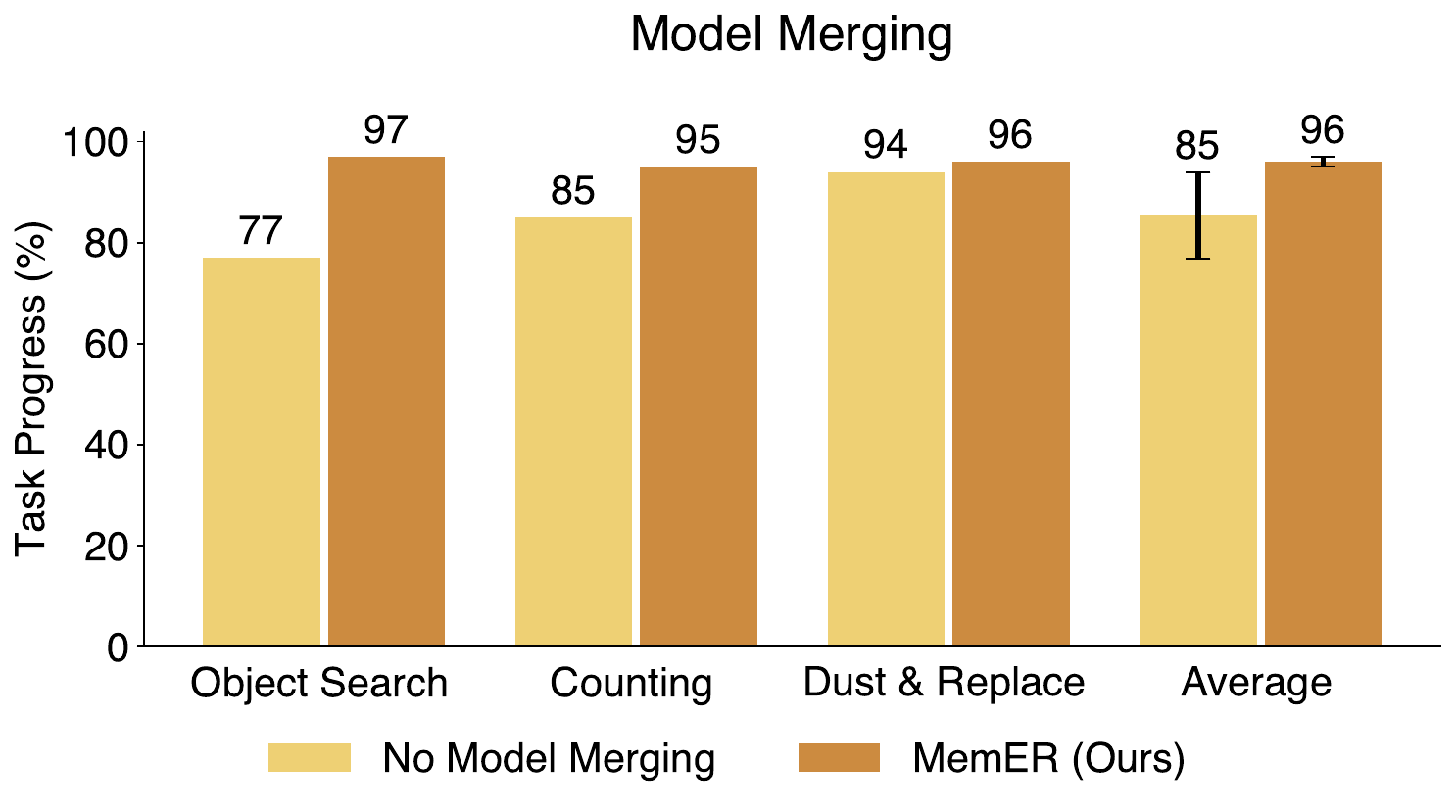}
\end{minipage}

\caption{\textbf{(Left) Modality Results.} Using only images to represent the memory performs better than the baselines that use only text or text and images. We hypothesize that the high-level policy over-indexes on the text tokens in the memory, causing it to miss important details in the visual input.
\textbf{(Right) Model Merging Results.} Merging the weights of our finetuned high-level policy with the pretrained Qwen2.5-VL-7B-Instruct weights helps or maintains performance with all tasks.}
\label{fig:modality_and_merging} %

\end{figure}

\textbf{Model Merging.} An important factor contributing to the success of our policy is the strong video understanding prior in Qwen2.5-VL-7B-Instruct \citep{qwen2.5}. However, training the high-level policy to accurately predict the language subtasks used by the low-level policy requires roughly 5,000 gradient steps (10–15 epochs). After this amount of finetuning, the high-level policy tends to lose some robustness to low-level policy freezes and retry behaviors, due to its training data consisting solely of optimal expert demonstrations. Concurrent work suggests that linearly interpolating the weights of a generalist pretrained model with those of the same model finetuned on narrow, task-specific data can help preserve the pretrained model’s robustness and generalization, while still allowing adaptation to the new task \citep{anon2025Robust}. We find this also applies to the high-level policy. Specifically, we set the weights of our high-level policy as: \begin{equation}
    \theta = (1 - \alpha)\cdot\theta_{\text{pre}} + \alpha \cdot \theta_{\text{ft}}
\end{equation}
where $\theta_{\text{pre}}$ is the weights of Qwen2.5-VL-7B-Instruct and $\theta_{\text{ft}}$ is the weights of this model finetuned on all three memory-based tasks. We follow \cite{anon2025Robust} and set $\alpha=0.8$ for all baselines we test. Figure \ref{fig:modality_and_merging} (right) shows that model merging improves or maintains performance across all tasks.

\textbf{Closed-Loop Deployment.} Our policy decomposition is the following:

\begin{equation}
      \pi(\boldsymbol{A_t}|o_{0:t}) = \pi_l(\boldsymbol{A_t} | \boldsymbol{I_t}, \boldsymbol{q_t}, \mathit{l_t'})\pi_h(l_t', \boldsymbol{J_t}|\boldsymbol{I_{t-N+1:t}}, \boldsymbol{K_t})
\end{equation}

The interaction between the low-level and high-level policy for closed loop deployment is shown in Figure \ref{fig:architecture}. The low-level policy predicts $\pi_l$ actions chunks at $\sim$2Hz, while the the high-level policy $\pi_h$ predicts keyframes and subtasks at roughly $\sim$1Hz. We run both policies on their own server. As in \cite{hirobot}, we choose to run the policies  asynchronously, as we find it to improve responsiveness and stability during deployment. While the high-level policy is predicting the next subtask, the low-level policy conditions on the latest predicted subtask. We add the image observations subsampled at 2Hz to a queue, and then send this queue to the high-level policy to query the next subtask prediction after the current high-level policy prediction is complete.

\section{Experiments}
In this section, we aim to evaluate the extent to which our method and alternative approaches can tackle long-horizon manipulation tasks that require memory. We first describe our tasks and evaluation protocols, then discuss the following questions:
\begin{enumerate}
    \item To what extent can our approach tackle tasks that require memory, in comparison to a memory-less policy (i.e. current robot foundation models), a human high-level (Human HL) policy, and other naive approaches?
    \item How does our high-level policy, finetuned from an open-source VLM, compare to proprietary off-the-shelf VLMs?
    \item How does representing memory via images compare to other modalities?
\end{enumerate}

\begin{figure}[t]
  \centering
  \includegraphics[width=\linewidth]{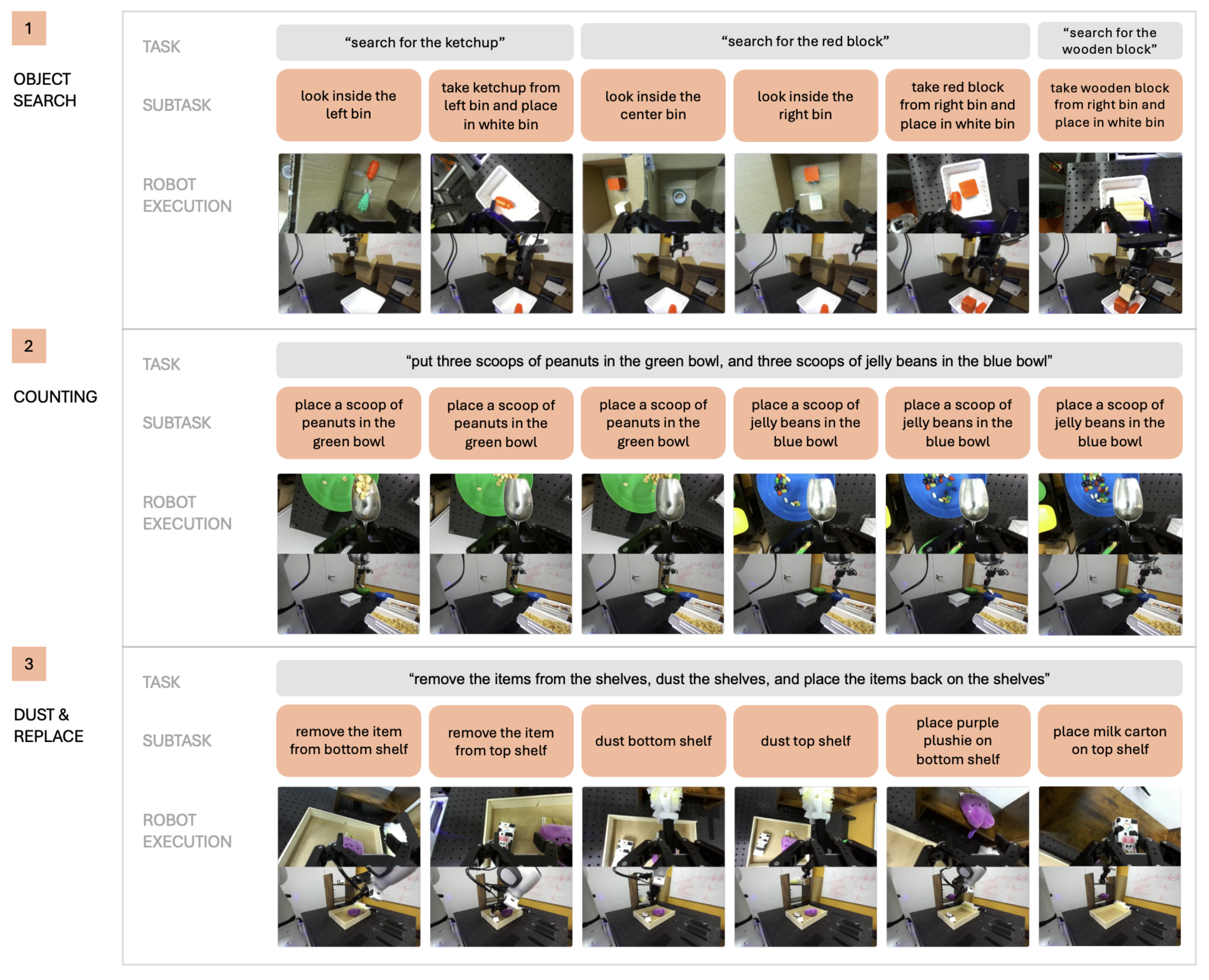}
  \caption{\textbf{Tasks used in our evaluations.} Across three domains, we evaluate complex instructions, intermediate subtasks, and keyframe predictions. We report performance across 20 trials per task per method.}
  \label{fig:tasksubtasks}
\end{figure}

\begin{figure}[t]
\centering
\begin{minipage}[b]{1.0\linewidth}
  \centering
  \includegraphics[width=\linewidth]{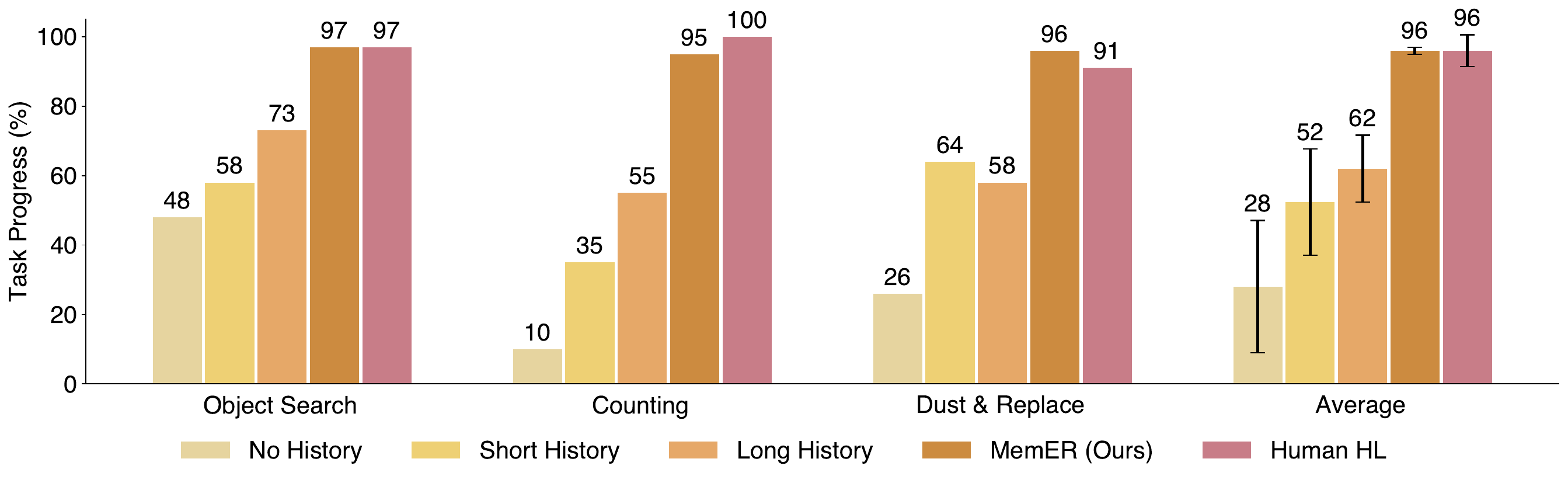}\hfill
\end{minipage}
\caption{\textbf{Main Results.} Our method clearly outperforms the no history, short history (8 frames of context), and long history (32 frames of context) baselines on the three long-horizon memory-based tasks. It is on par with the human high level policy. }
\label{fig:main_results}
\end{figure}

We design three complex, real-robot tasks that entail using memory in multiple distinct ways, including remembering object locations, keeping track of previously completed actions, and counting repeated steps, as illustrated in Fig. \ref{fig:tasksubtasks}. Since all of the tasks are long-horizon, we record different metrics for each task to provide a granular view of task completion.

\textbf{Object Search.} In this task, we randomly place three to five objects across three opaque bins. Then, the robot is sequentially given three objects to find; each new instruction is issued only after the robot has attempted to retrieve the previous object. Our goal is an optimized search: the robot remembers which bins it has already examined (and what it saw), skips re-searching them, and explores additional bins only as needed. It should proceed directly to the target bin if it has already looked inside it before. This task tests cross-episodic memory as finding each object requires its own $l_t$, thus necessitating recall of information gathered while executing prior instructions. We train and test with the same set of 15 objects, which are various small toys (see Figure \ref{fig:all_task_objects} in Appendix \ref{sec:cross-task-objs}). \textit{Evaluation metric.} We measure task completion by two criteria for each of the three objects: successful retrieval and adherence to the optimal path without unnecessary exploration, for a maximum score of 6 (2 points per object). We report results in Table \ref{tab:bin-counting-dusting-q1}, along with the overall accuracy in Figure \ref{fig:main_results}.

\textbf{Counting Scoops.} In this task, the robot is asked to place an exact number of scoops of two different ingredients into two different bowls. The robot needs to keep track of how many scoops have already been obtained per ingredient. This counting task has appeared in prior work \citep{ptp}, and we modify it to require much longer-horizon reasoning by increasing the number of scoops and ingredients to scoop from and varying the number of scoops across instructions. This task is challenging because the keyframes corresponding to each ingredient are nearly indistinguishable, so missed or duplicated keyframes cause the high-level policy to misjudge its progress. We train and test with peanuts and jelly beans (see Figure \ref{fig:all_task_objects} in Appendix \ref{sec:cross-task-objs}).  %
\textit{Evaluation metric.} Task completion is measured by the absolute value of the difference between the number of scoops requested and obtained for each ingredient, so a lower metric is better. We report results in Table \ref{tab:bin-counting-dusting-q1}, along with the 0-1 success rate for satisfying the instruction in Figure \ref{fig:main_results}.

\textbf{Dust \& Replace.} In this task, the robot is asked to remove objects from a two-tiered shelf, pick up a duster, dust each shelf, and replace the objects to their original positions. Between dusting the two shelves, we return the duster to an ambiguous position that makes it unclear from recent context which shelf has already been dusted. This task is challenging because the robot must simultaneously remember two types of information: the original locations of the objects and which shelf, if any, has already been dusted. We train on a set of 17 objects and test with a subset of 9 objects from the training set, which are various plushies. We removed some from objects when testing since the low-level policy could not manipulate them, and we are primarily evaluating the capabilities of the high-level policy (see Figure \ref{fig:all_task_objects} in Appendix \ref{sec:cross-task-objs}). 
\textit{Evaluation metric.} Task completion is measured by the binary success of each object being replaced correctly on the shelf and each shelf being dusted, for a maximum score of 4. We report results in Table \ref{tab:bin-counting-dusting-q1} and the overall task accuracy in Figure \ref{fig:main_results}.

\textbf{Evaluation Setup:} Our robot setup resembles what was used in DROID \citep{khazatsky2025droidlargescaleinthewildrobot}, with a Franka arm, parallel jaw gripper and two cameras (third-person ZED camera and a wrist-mounted miniZED camera). For all tasks, the $\pi_h$ operates at $\sim$1Hz and the $\pi_l$ operates at $\sim$2Hz. $\pi_l$ outputs an action chunk $\boldsymbol{A_t}$ of 15 actions sampled at 15Hz, and we execute 8 actions open-loop before replanning. The cameras stream $320 \times 180$ resolution images at 15Hz, but we subsample to 2Hz as input to our hierarchical policy.

\begin{table}[!t]
\centering
\footnotesize 
\setlength{\tabcolsep}{3pt} %
\renewcommand{\arraystretch}{1.1} %

\begin{tabular*}{\linewidth}{@{\extracolsep{\fill}}lcccccccc@{}}
\toprule
\textbf{Method} 
& \multicolumn{2}{c}{\textbf{Object Search}} 
& \multicolumn{1}{c}{\textbf{Counting}} 
& \multicolumn{4}{c}{\textbf{Dust \& Replace}} \\
\cmidrule(lr){2-3} \cmidrule(lr){4-4} \cmidrule(l){5-8}

& \shortstack{\textbf{\# times} \\ \textbf{object} \\ \textbf{retrieved} (\(\uparrow\))} 
& \shortstack{\textbf{\# times used} \\ \textbf{optimal} \\ \textbf{path} (\(\uparrow\))} 
& \shortstack{\textbf{\# wrong} \\ \textbf{scoops} \\ (\(\downarrow\))}
& \shortstack{\textbf{Dust} \\ \textbf{bottom} \\ \textbf{shelf} (\(\uparrow\))}
& \shortstack{\textbf{Dust} \\ \textbf{top} \\ \textbf{shelf} (\(\uparrow\))}
& \shortstack{\textbf{Replace} \\ \textbf{bottom} \\ \textbf{object} (\(\uparrow\))}
& \shortstack{\textbf{Replace} \\ \textbf{top} \\ \textbf{object} (\(\uparrow\))} \\
\midrule
\fullname\space (Ours)   & \textbf{59} & \textbf{57} & \textbf{1} & \textbf{20} & \textbf{19} & \textbf{18} & \textbf{20} \\
No history  & 32 & 25 & 61 & 5 & 4 & 5 & 7 \\
Short History & 38 & 31 & 26 & 14 & 14 & 11 & 12 \\
Long History & 47 & 41 & 12 & 11 & 11 & 12 & 12 \\
\midrule
Human HL    & 58 & 58 & 0 & 19 & 19 & 18 & 17 \\
\bottomrule
\end{tabular*}
\caption{\textbf{Detailed Main Results.} Online evaluation of our method and the baselines for Q1. We provide task-specific evaluation metrics and the raw counts across 20 trials for each component of the task. Bold marks the best non-oracle method in each row. \(\uparrow\) and \(\downarrow\) indicate higher and lower is better, respectively.}
\label{tab:bin-counting-dusting-q1}
\vspace{-0.4em}
\end{table}

\subsection{Main Results}
\textbf{Q1: To what extent can our approach tackle tasks that require memory compared to other methods?}
All evaluated methods incorporate a $\pi_h$ and $\pi_l$, and the baselines change the input context of the high-level policy $\pi_h$ while using the same $\pi_l$. We compare to the following baselines: 1) No history: a memory-less high-level policy that only views the current frame (i.e. current robot foundation models), similar to \citep{hirobot} 2) Short History: a policy that views only the recent $N$ frames ($N=8$ for our setup) 3) Long History: a policy that naively receives a longer context (4$\times$ that of Short History or $N=32$ recent frames), and 4) Human HL: a human provides the correct subtasks. The Human HL policy establishes a rough estimate of the upper bound performance for all tasks, with failures stemming from the low-level policy. 
From Figure \ref{fig:main_results}, we see that No History and Short History baselines perform poorly as all of the tasks simply require more context than what was provided. The Long History baseline shows that increasing the context can slightly help, but 32 frames ($\sim$16 seconds of memory) incurs an inference cost of 1 second, which approaches the limit of what can be tolerated in closed-loop settings. 
The Long History policy still performs on average $34\%$ worse than \fullname, necessitating strategies such as our method that consolidate keyframes rather than naively extending the context. 
Lastly, our method has $>90\%$ on all tasks with the most common failure case being failures in the low-level policy executing the subtask, which can be rectified with better low-level correction data.

\textbf{Q2: How does our high-level policy, finetuned from an open-source VLM, compare to proprietary off-the-shelf VLMs?}
Since our approach outperforms other selections of $\pi_h$ and performs similar to the Human HL policy, we investigate if our method is necessary given existing state-of-the-art VLMs may already have this capability. We test both GPT-5 and Gemini Robotics–ER 1.5 given the former's strong multimodal reasoning performance and the latter's robotics-specific agentic capabilities. Because the API latency for both ranged from 10-15 seconds, these API-based high-level policies led to complete failures when we deployed it in the same closed-loop evaluation as the other baselines, which require latencies of less than 1 second to react accordingly to the environment. 

To still offer a means of comparison between $\pi_h$ and the API-based high-level policies, we designed an offline experiment using a held-out set of trajectories generated by the low-level policy commanded by ground-truth subtasks $l_t'$. This simulates closed-loop execution under realistic behaviors (i.e. retries after missed grasps, pauses, and jerky motions), while allowing the model to build its visual memory in the same way. We carefully craft the prompt to include specific task-relevant instructions and an explicit list of all possible actions that the low-level policy can follow, and ask the model to choose among them (Appendix \ref{sec:gpt_prompt}). 
Just like our setup, the model takes in the $N=8$ most recent frames of context and selected keyframes $\boldsymbol{K_t}$ at every timestep, and outputs the subtask $l'_t$ for the low-level policy to execute and candidate keyframes $\boldsymbol{J_t}$. We measure \textit{trajectory accuracy}, which is how often the correct subtask is predicted at each timestep in the trajectory, since we know the ground-truth subtask command that the low-level policy is executing at that moment. We also measure \textit{boundary accuracy}, computed as the trajectory accuracy within a fixed window centered on transition points between subtasks. These are critical moments that expose the high-level policy’s grasp of task progress by knowing when to move on to the next subtask; correct timing in transitioning between subtasks plays a major role in proper coordination with the low-level policy during deployment. From Table \ref{tab:subtask-success-boundary}, we observe that both zero-shot API-based models perform poorly compared to our finetuned Qwen2.5-VL model, primarily failing by predicting too many non-informative candidate keyframes, reflecting its limited ability to identify which frames are truly useful. Consequently, even with a significantly stronger base VLM such as GPT-5 or Gemini Robotics–ER 1.5, the model lacks the capacity to interpret robot-specific perceptual cues and identify meaningful keyframes, resulting in less accurate subtask predictions and necessitating additional fine-tuning.

\begin{table}[!t]
\centering
\footnotesize %
\setlength{\tabcolsep}{5pt} %
\renewcommand{\arraystretch}{0.95}
\begin{tabular}{@{}l*{3}{cc}@{}}
\toprule
\textbf{Method}
& \multicolumn{2}{c}{\textbf{Object Search}}
& \multicolumn{2}{c}{\textbf{Counting}}
& \multicolumn{2}{c}{\textbf{Dust \& Replace}} \\
\cmidrule(lr){2-3}\cmidrule(lr){4-5}\cmidrule(l){6-7}
& \shortstack{\textbf{Trajectory} \\ \textbf{acc.} (\(\uparrow\))} & \shortstack{\textbf{Boundary} \\ \textbf{acc.} (\(\uparrow\))}
& \shortstack{\textbf{Trajectory} \\ \textbf{acc.} (\(\uparrow\))} & \shortstack{\textbf{Boundary} \\ \textbf{acc.} (\(\uparrow\))}
& \shortstack{\textbf{Trajectory} \\ \textbf{acc.} (\(\uparrow\))} & \shortstack{\textbf{Boundary} \\ \textbf{acc.} (\(\uparrow\))} \\
\midrule
\fullname\space (Ours) & \textbf{0.80} & \textbf{0.76} & \textbf{0.67} & \textbf{0.65} & \textbf{0.87} & \textbf{0.86} \\
GPT\textendash 5 & 0.15 & 0.16 & 0.43 & 0.47 & 0.67 & 0.63 \\
Gemini Robotics\textendash ER 1.5 & 0.21 & 0.23 & 0.13 & 0.14 & 0.19 & 0.22 \\
\bottomrule
\end{tabular}

\caption{\textbf{Comparison with API-Based VLMs.} Offline evaluations of the per-task trajectory and boundary accuracy of subtask predictions between \fullname, GPT-5, and Gemini Robotics-ER 1.5, to compare our finetuned high-level policy from an open-source VLM against proprietary VLMs.}
\label{tab:subtask-success-boundary}
\end{table}

\begin{table}[!t]
\centering
\renewcommand{\arraystretch}{0.95}
\resizebox{\columnwidth}{!}{%
\begin{tabular}{@{}l ccc ccccccc@{}}
\toprule
\textbf{Method} 
& \multicolumn{3}{c}{\textbf{Input Components}} 
& \multicolumn{2}{c}{\textbf{Object Search}} 
& \multicolumn{1}{c}{\textbf{Counting}} 
& \multicolumn{4}{c}{\textbf{Dust \& Replace}} \\
\cmidrule(lr){2-4} \cmidrule(lr){5-6} \cmidrule(lr){7-7} \cmidrule(l){8-11}

& \shortstack{\textbf{Short} \\ \textbf{History}} 
& \shortstack{\textbf{Image} \\ \textbf{Keyframes}} 
& \shortstack{\textbf{Text} \\ \textbf{Subtasks}} 
& \shortstack{\textbf{\# times} \\ \textbf{object} \\ \textbf{retrieved} (\(\uparrow\))} 
& \shortstack{\textbf{\# times used} \\ \textbf{optimal} \\ \textbf{path} (\(\uparrow\))} 
& \shortstack{\textbf{\# wrong} \\ \textbf{scoops} \\ (\(\downarrow\))} 
& \shortstack{\textbf{Dust} \\ \textbf{bottom} \\ \textbf{shelf} (\(\uparrow\))} 
& \shortstack{\textbf{Dust} \\ \textbf{top} \\ \textbf{shelf} (\(\uparrow\))} 
& \shortstack{\textbf{Replace} \\ \textbf{bottom} \\ \textbf{object} (\(\uparrow\))} 
& \shortstack{\textbf{Replace} \\ \textbf{top} \\ \textbf{object} (\(\uparrow\))} \\
\midrule

\fullname\space (Ours) & $\checkmark$ & $\checkmark$ & $\times$ & \textbf{59} & \textbf{57} & \textbf{1} & \textbf{20} & \textbf{19} & \textbf{18} & \textbf{20} \\
\shortstack{Short History + Text} & $\checkmark$ & $\times$ & $\checkmark$ & 40 & 28 & 10 & 16 & 16 & 7 & 10 \\
\shortstack{MemER + Text} & $\checkmark$ & $\checkmark$ & $\checkmark$ & \textbf{59} & 49 & 13 & \textbf{20} & 18 & 17 & \textbf{20} \\
\bottomrule
\end{tabular}%
} %
\caption{\textbf{Detailed Modality Results.} Online evaluation across methods ablating the textual modality. Bold marks the best method. \(\uparrow\) and \(\downarrow\) indicate higher and lower is better, respectively.}
\label{tab:bin-counting-dusting-q3}
\end{table}

\textbf{Q3: How does representing memory via images compare to other modalities?}
We now discuss which modalities are best suited for building memory—visual, textual, or both. Storing memory in text offers natural benefits as it's interpretable and much more condensed. We test two additional methods that use text memory, in the form of the predicted subtask $l'_t$ that is associated with each of the selected keyframes in $\boldsymbol{K_t}$: 1) Short History + Text uses the most recent $N=8$ frames and predicted subtasks and 2) \fullname\space + Text interleaves the predicted subtasks and visual keyframes in memory. 
Table \ref{tab:bin-counting-dusting-q3} shows the input for each baseline.

We see from Figure \ref{fig:modality_and_merging} (left) that relying on textual memory underperforms compared to our vision-only approach. Specifically, replacing the visual memory with text (Short History + Text) leads to the most significant performance drop. Furthermore, adding text to our visual memory (\fullname\space+ Text) provides no benefits, consistently under-performing across all tasks, especially the Counting task. We find that both the baselines' subtask predictions are brittle, largely due to overreliance on the most recently predicted subtask stored in memory. This leads to failures when policy retries or freezes shift the recent context out of distribution. In such cases, the model tends to overfit to the canonical ordering of subtasks observed in expert demonstrations and misidentifies the subtask being executed given the current environmental state. In contrast, directly grounding predictions in the current observation combined with the robust visual memory proves more reliable. 

For the Short History + Text baseline, the language-based subtasks do not capture all of the information required to successfully complete the task. For example, in the Object Search task, the predicted language subtasks only specify the objects the robot has previously been asked to locate or is currently searching for, but have no reference to objects it has seen that may need to be retrieved in subsequent episodes. For the \fullname\space + Text baseline, the model disproportionately attends to the text stored in memory, which can be incorrect for the reasons stated above, and subsequently ignores important information stored in visual memory. Such behavior has been noted before in \citep{zheng2025mllmsdeeplyaffectedmodality, lee2025vlindbenchmeasuringlanguagepriors}. 
Thus, from our tasks, we find that visual memory alone provides the most robust representation, though exploring multimodal memory remains an interesting future direction.

\section{Discussion and Future Work}
We introduced \fullname, a hierarchical vision–language–action framework that \emph{scales memory via experience retrieval}. A high-level memory policy processes streamed observations, nominates keyframes to retain, and emits language subtasks that a low-level controller executes. A simple online consolidation strategy converts per-timestep candidate keyframes into a compact, stable episodic memory that is fed back into the high-level policy. Across three real-world, long-horizon manipulation domains, \fullname{} significantly improves performance on tasks requiring minutes of recall while retaining low-latency inference and strong compatibility with existing VLA backbones.

Despite its benefits, our approach has several limitations. %
We continuously accumulate informative keyframes but currently lack a mechanism to discard them when they become too numerous—an issue that may arise for tasks requiring hours of memory. Enabling the high-level policy to reason about which keyframes to not only \textit{add} but also \textit{delete} for modifiable long-term memory is an exciting direction for future work. Our throughput also depends on the VLM backbone and on scheduling (e.g., $\pi_h$ at 1 Hz, $\pi_l$ at 2 Hz), which can limit very high-frequency control and reduce reactivity to rapid environmental changes. Incorporating improved model caching and better tokenization could further lower inference latency, enabling higher-frequency control. In addition, our memory is limited to visual observations; incorporating other sensory modalities such as tactile or audio is a promising extension. Finally, we study a single robot embodiment, and extending to mobile manipulation and multi-room tasks, where memory must interleave spatial mapping with episodic recall, would bring the system closer to human-like memory. We view \fullname{} as a step toward robot policies that \emph{decide what to remember} and \emph{leverage those memories when needed} for effective long-horizon control.

\section{Acknowledgments}
This research was supported by ONR grant N00014-22-1-2621, the Robotics and AI Institute, and compute from the Stanford Institute for Human-Centered AI (HAI). We thank Dorsa Sadigh, John Yao, Moo Jin Kim, Lucy Xiaoyang Shi, Brian Kim, Homer Walke, Marcel Torne, Yuejiang Liu, Anikait Singh, Andy Tang, and Jennifer Grannen for their informative discussions. We thank Yuejiang Liu, John Yao, Alex Swerdlow, and Ria Doshi for their feedback on earlier versions of the paper. We also thank Physical Intelligence for providing beta access to the $\pi_{0.5}$ model. Ajay Sridhar is supported by the NSF Graduate Research Fellowship.

\bibliography{references}
\bibliographystyle{iclr2026_conference}

\clearpage
\appendix
\section{Model Initialization and Hyperparameters}
\label{sec:finetuning_hyperparams}

\textbf{Training the High-Level Policy} The hyperparameters for fine-tuning the Qwen2.5-VL-7B-Instruct high-level policy are detailed in Table \ref{tab:high_level_hyperparams}.

\begin{table}[h]
\centering
\caption{Hyperparameters for High-Level Policy (Qwen2.5-VL-7B-Instruct) fine-tuning.}
\label{tab:high_level_hyperparams}
\begin{tabular}{ll}
\toprule
Hyperparameter & Value \\
\midrule
Optimizer & AdamW \\
$\beta_{1}$ & 0.9 \\
$\beta_{2}$ & 0.999 \\
Weight Decay & 0 \\
Gradient Clip Norm & 1.0 \\
LR Schedule & Cosine \\
Warmup Ratio & 0.05 \\
Batch Size & 256 \\
Training & 4500 gradient steps \\
Compute & 96 H200 GPU hours \\
Frozen Layers & Vision Encoder, Projection Layer \\
Trainable Layers & LLM Backbone \\
\bottomrule
\end{tabular}
\end{table}
\FloatBarrier

\textbf{Training the Low-Level Policy} The hyperparameters for fine-tuning the $\pi_{0.5}$ low-level policy are detailed in Table \ref{tab:low_level_hyperparams}. The model is fine-tuned from the public $\pi_{0.5}$ checkpoint trained on the DROID dataset \citep{khazatsky2025droidlargescaleinthewildrobot}.

\begin{table}[h]
\centering
\caption{Hyperparameters for Low-Level Policy ($\pi_{0.5}$) fine-tuning.}
\label{tab:low_level_hyperparams}
\begin{tabular}{ll}
\toprule
Hyperparameter       & Value                 \\
\midrule
Optimizer            & AdamW                 \\
$\beta_{1}$          & 0.9                   \\
$\beta_{2}$          & 0.95                  \\
Weight Decay         & 0                     \\
Gradient Clip Norm   & 1.0                   \\
LR Schedule          & Cosine                \\
Warmup Ratio         & 0.033                 \\
Batch Size           & 128                   \\
Training Steps       & 18000                 \\
Compute              & 48 H200 GPU hours     \\
\bottomrule
\end{tabular}
\end{table}
\FloatBarrier

\section{Data Collection and Labeling the Subtasks}

\begin{figure}[htb!]
  \centering
  \includegraphics[width=0.75\linewidth]{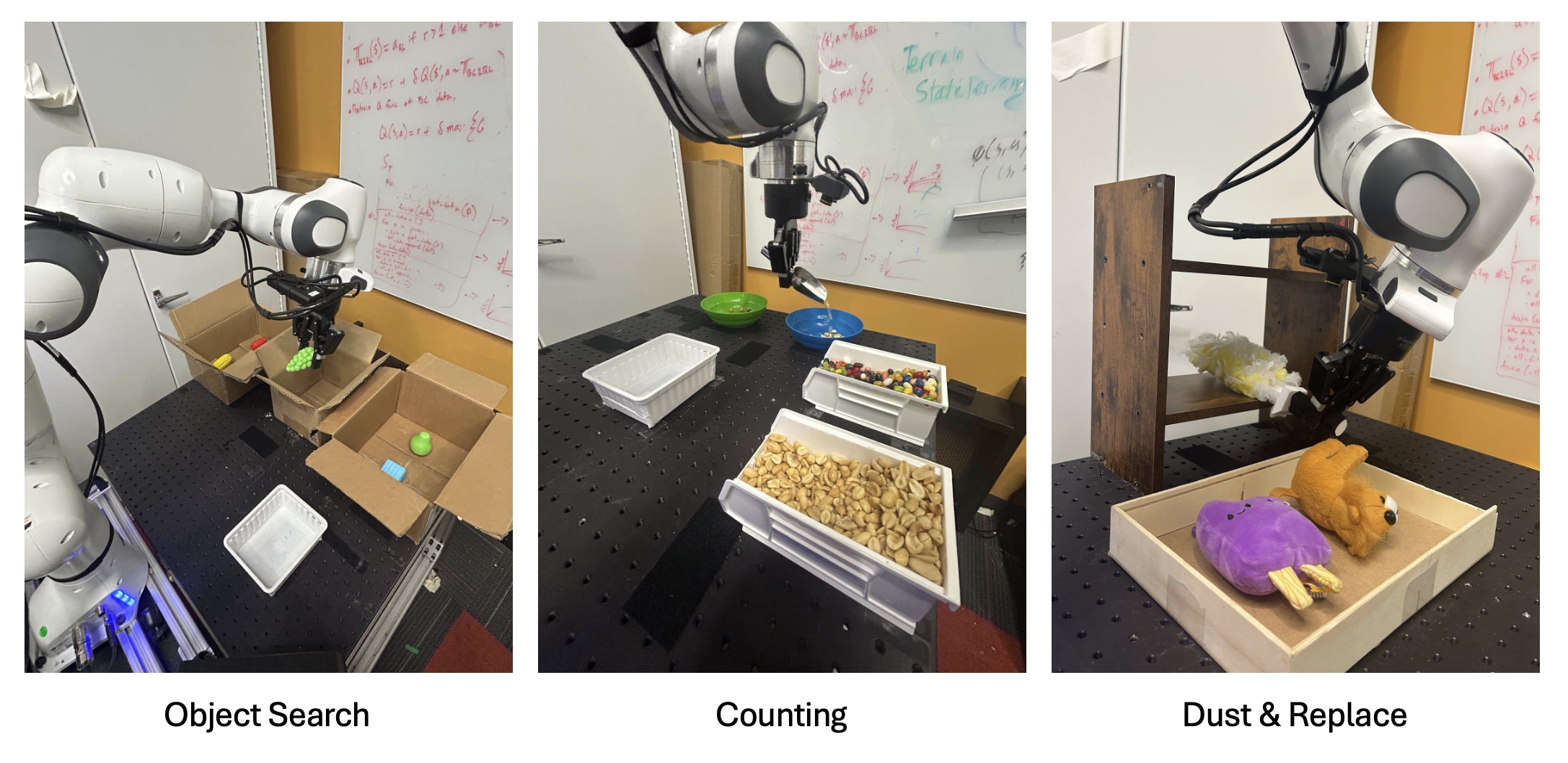}
  \label{fig:task}
    \caption{Images of the three tasks. Videos can be viewed at \website.}
\end{figure}

\label{app:task}
To collect the robot trajectory data, we follow the data collection procedure with the Oculus teloperation set in \cite{khazatsky2025droidlargescaleinthewildrobot}. To make the subtask labeling process as easy as possible, we generate a list of subtasks for the whole task trajectory prior to data collection. While collecting the data, we just need to follow the current subtask, and indicate when a subtask has been finished with a simple keyboard input. We automate the randomization of the high-level task to avoid human biases when collecting data.

\section{Cross-Task Object Generalization}
\label{sec:cross-task-objs}

\begin{figure}[t]
\centering

\begin{minipage}[t]{0.49\linewidth}
  \centering
  \includegraphics[width=\linewidth]{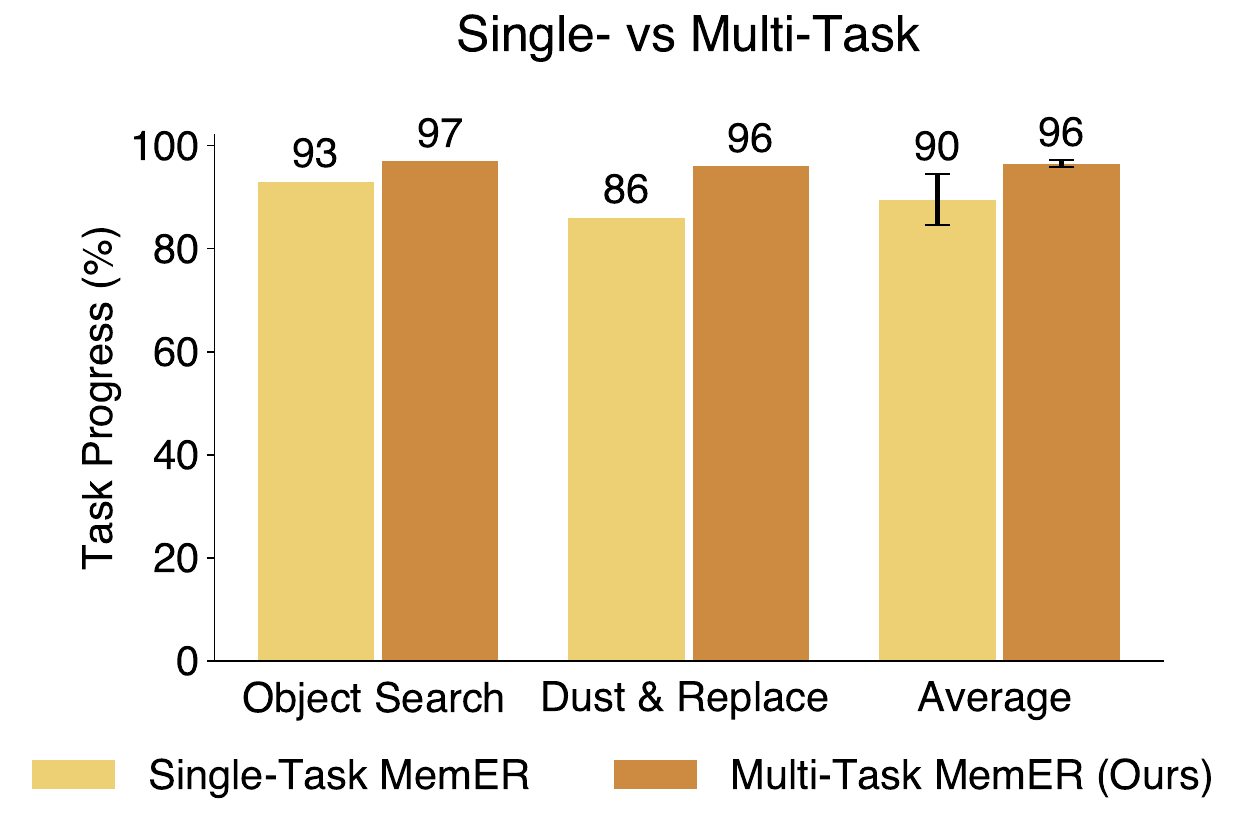}
\end{minipage}
\hfill %
\begin{minipage}[t]{0.48\linewidth}
  \centering
  \includegraphics[width=\linewidth]{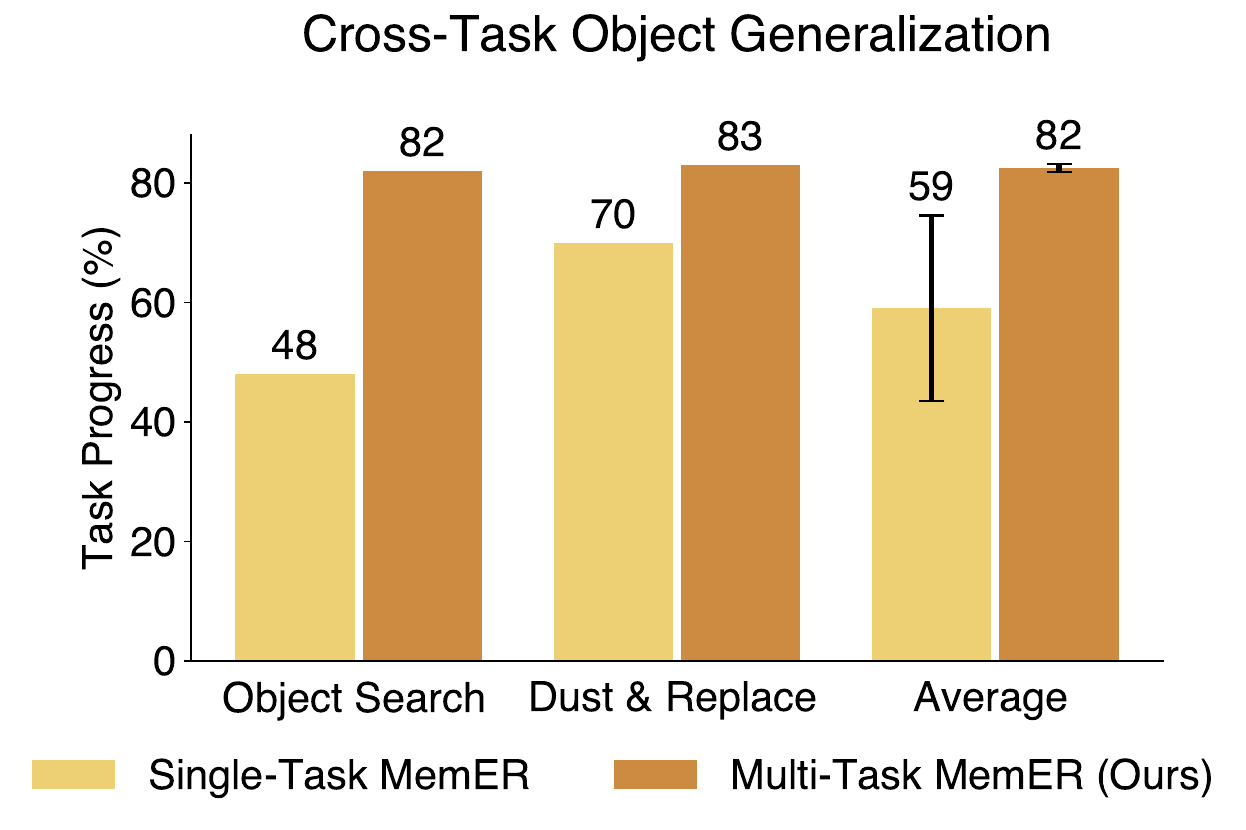}
\end{minipage}

\caption{\textbf{(Left) Single- vs Multi-Task Results.} The performance of the single- and multi-task versions of \fullname\space on the Object Search and Dust \& Replace task are similar. \textbf{(Right) Cross-Task Object Generalization Results.}  We swap objects between the Object Search and Dust \& Replace (see Figure \ref{fig:all_task_objects}) tasks during evaluation. The multi-task policy can generalize to the new object-task combinations during evaluation despite never seeing them in training.}

\label{fig:single_vs_multi} %

\end{figure}

\begin{figure}[tb]
\centering
\begin{minipage}[t]{0.22\linewidth}
  \centering
  \includegraphics[width=\linewidth, angle=90]{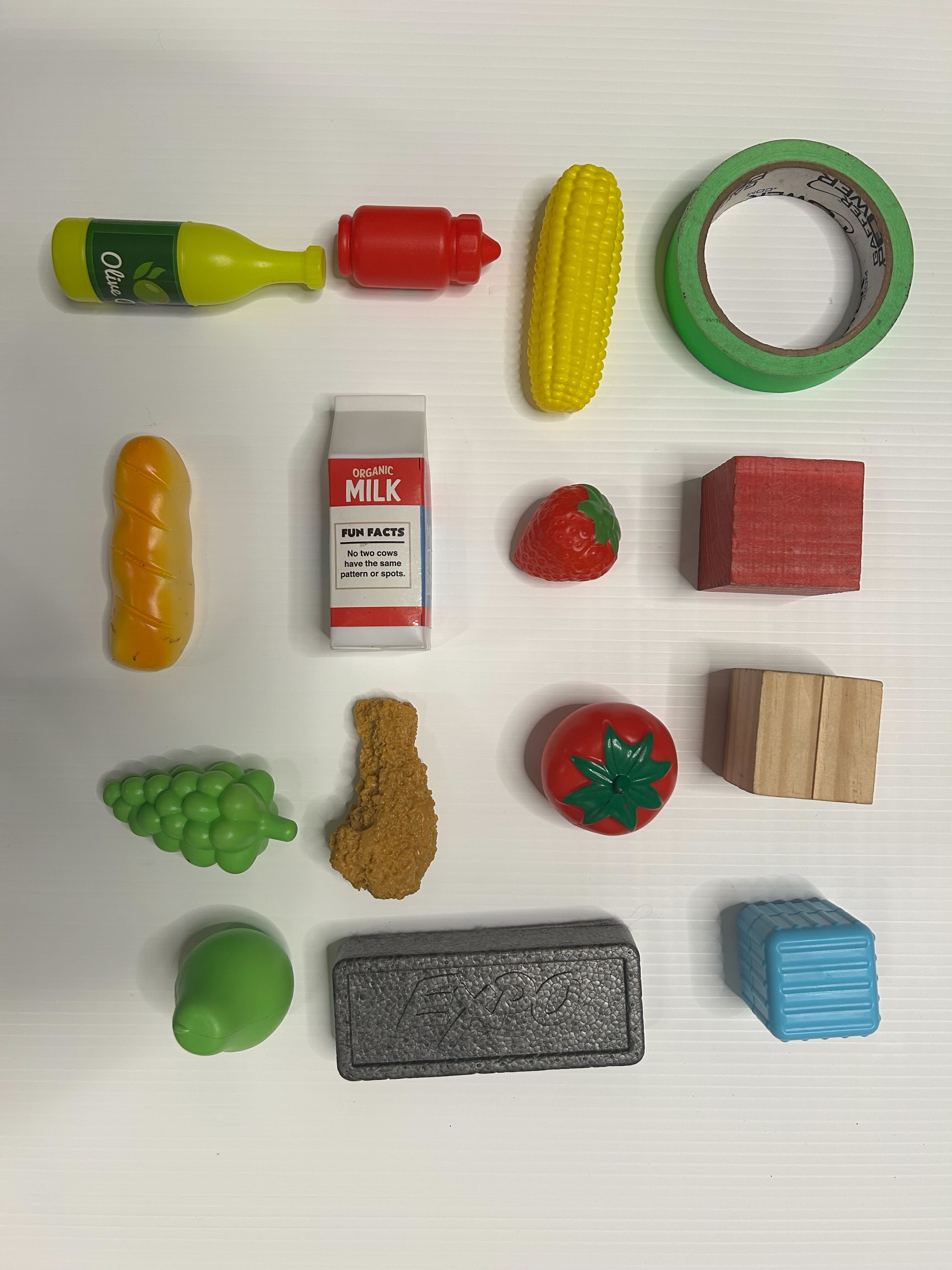}
\end{minipage}%
\hspace{0.08\linewidth} %
\begin{minipage}[t]{0.22\linewidth}
  \centering
  \includegraphics[width=\linewidth, angle=90]{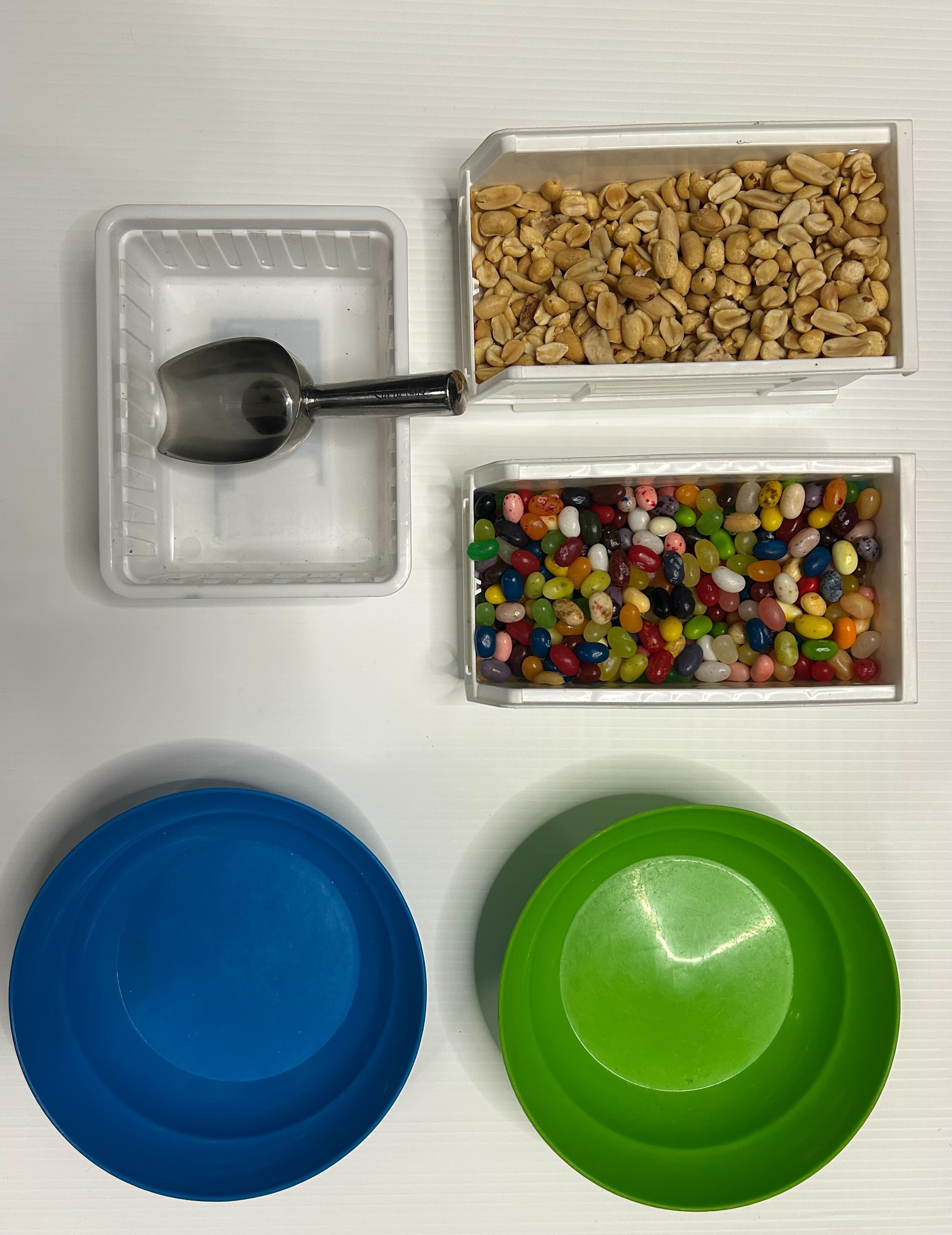}
\end{minipage}%
\hspace{0.08\linewidth} %
\begin{minipage}[t]{0.22\linewidth}
  \centering
  \includegraphics[width=\linewidth, angle=90]{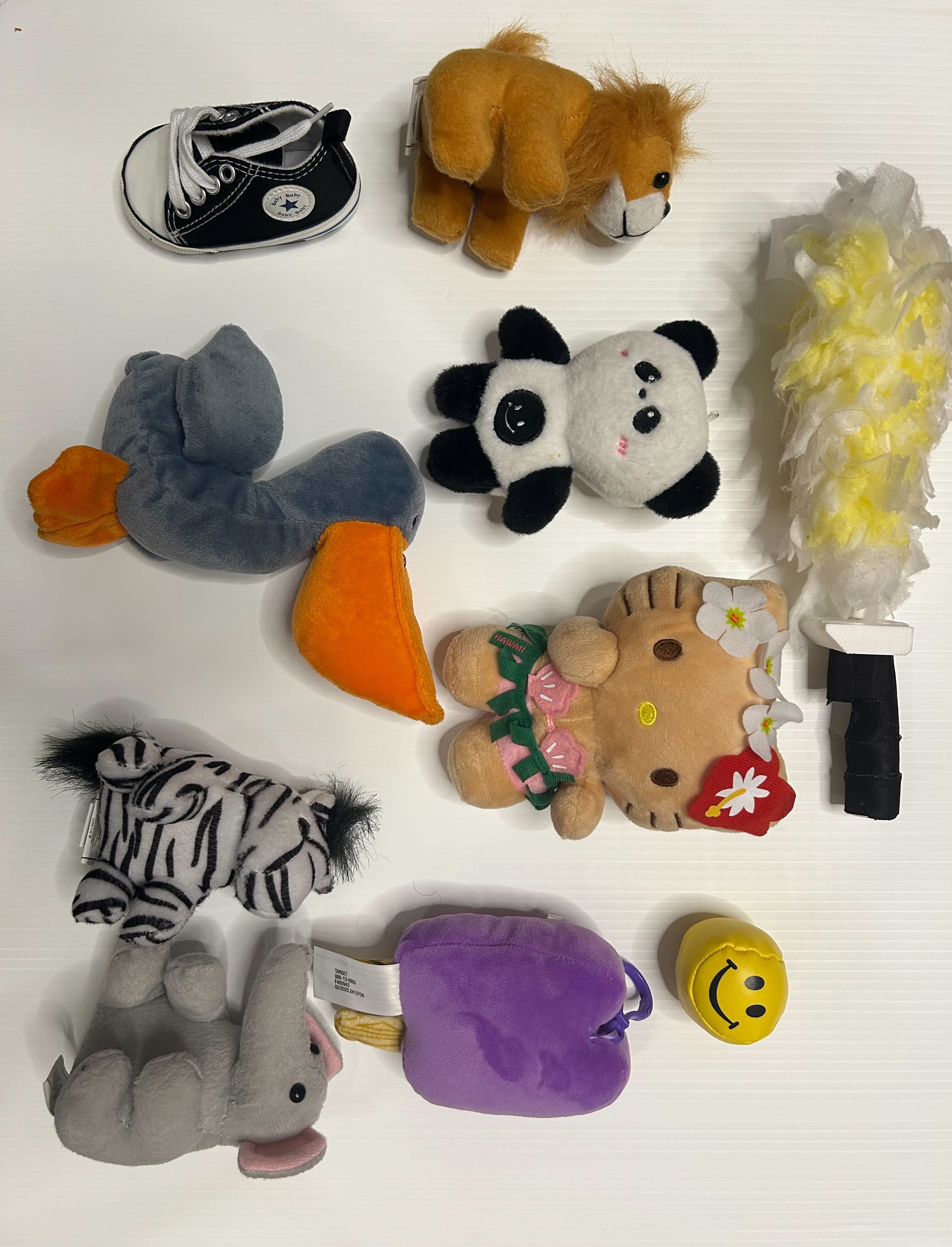}
\end{minipage}

\caption{\textbf{(Left)} Objects for the Object Search task. 
\textbf{(Center)} Objects for the Counting task. 
\textbf{(Right)} Objects for the Dust \& Replace task.}
\label{fig:all_task_objects} %

\end{figure}

To evaluate the benefits of multi-task training, we compare a single-task version of \fullname\space (separate policy trained for each task) with our multi-task version. We evaluate on our object-centric tasks: the Object Search and Dust \& Replace tasks, with objects from Figure  \ref{fig:all_task_objects}. First, we establish a baseline by evaluating both versions of \fullname\space on their original tasks. As shown in Figure \ref{fig:single_vs_multi} (left), their performance is roughly similar. 

The primary benefit of multi-task training is revealed when evaluating cross-task object generalization. For this evaluation, we swap all of the objects between the tasks (e.g., using Object Search objects for the Dust \& Replace task, and vice-versa). This creates new object-task combinations that the models have not seen during training. The results in Figure \ref{fig:single_vs_multi} (right) demonstrate the clear advantage of using the multi-task model, generalizing much more effectively to new object-task combinations than the single-task version. This shows that our method's ability to generalize improves by scaling the number of memory-based tasks we train on.

\section{Keyframe Selection Algorithm}
\label{app:algo}

\begin{algorithm}[H]
\caption{Selecting Keyframes from Candidates}\label{alg:build_memory}
\begin{algorithmic}[1]
    \Require 
    \begin{tabular}[t]{@{}l@{}}
        A sequence of candidate keyframe sets $\boldsymbol{J'_{0:t}} = (\boldsymbol{J_0},\boldsymbol{J_1}, \ldots, \boldsymbol{J_t})$ \\
        The merge distance $d$
    \end{tabular}
    \Ensure A list of the selected keyframes $\boldsymbol{K_t}$
    \Function{BuildVisualMemory}{$\boldsymbol{J'}_{0:t}$, $d$}
        \State $\boldsymbol{G_{0:t}} \gets \text{Sort}(\text{GetIndicesFromFrames}(\boldsymbol{J'}_{0:t}))$ \Comment{Extract the temporal indices.}
        
        \If{$\boldsymbol{G_{0:t}}$ is empty} \Comment{Handle case with no candidates.}
            \State \Return $\emptyset$
        \EndIf
        
        \State $Clusters \gets []$
        \State $C_{current} \gets [\boldsymbol{G_{0:t}}[0]]$
        
        \For{$i = 1$ to $|\boldsymbol{G_{0:t}}|-1$} \Comment{Build the clusters.}
            \If{$\boldsymbol{G_{0:t}}[i] - \boldsymbol{G_{0:t}}[i-1] \leq d$}
                \State Append $\boldsymbol{G_{0:t}}[i]$ to $C_{current}$
            \Else
                \State Append $C_{current}$ to $Clusters$
                \State $C_{current} \gets [\boldsymbol{G_{0:t}}[i]]$ \Comment{Start a new cluster.}
            \EndIf
        \EndFor
        \State Append $C_{current}$ to $Clusters$

        \State $T_{selected} \gets []$ 
        \For{each cluster $C$ in $Clusters$} \Comment{Select the median index of each cluster.}
            \State $i_{median} \gets \text{Median}(C)$
            \State Append $i_{median}$ to $T_{selected}$
        \EndFor
        
        \State $\boldsymbol{K_t} \gets \text{GetFramesFromIndices}(T_{selected})$
        \State \Return $\boldsymbol{K_t}$
    \EndFunction
\end{algorithmic}
\end{algorithm}

\section{Annotation Rules for Keyframes}
\label{sec:appendix_keyframe_rules}

As described in Section \ref{sec:pratical_implementation}, we use a simple, semantic annotation rule for each subtask interval to build the set of keyframes that constitute the ground-truth targets. The rules for each of the three tasks are as follows:

\begin{itemize}
    \item \textbf{Object Search}
    \begin{itemize}
        \item Select the \textbf{last frame} of \texttt{"look inside the <LOCATION> bin"}.
        \item Select \textbf{no frames} from \texttt{"take the <OBJECT> from the <LOCATION> bin and place it in the white bin"}.
    \end{itemize}
    
    \item \textbf{Counting}
    \begin{itemize}
        \item Select \textbf{no frames} from \texttt{"pick up the scooper"}.
        \item Select the \textbf{last frame} of \texttt{"place a scoop of <OBJECT> in the <COLOR> bowl"}.
        \item Select \textbf{no frames} from \texttt{"reset scooper position"}.
        \item Select \textbf{no frames} from \texttt{"drop the scooper"}.
    \end{itemize}

    \item \textbf{Dust \& Replace}
    \begin{itemize}
        \item Select the \textbf{last frame} from \texttt{"remove the object on the bottom shelf"}.
        \item Select the \textbf{last frame} of \texttt{"remove the object on the top shelf"}.
        \item Select the \textbf{last frame} of \texttt{"dust bottom shelf"}.
        \item Select \textbf{no frames} from \texttt{"reset duster"}.
        \item Select the \textbf{last frame} of \texttt{"dust top shelf"}.
        \item Select \textbf{no frames} from \texttt{"put down duster"}.
        \item Select the \textbf{last frame} from \texttt{"place the <OBJECT> on the bottom shelf"}.
        \item Select the \textbf{last frame} from \texttt{"place the <OBJECT> on the top shelf"}.
    \end{itemize}
\end{itemize}

\section{Prompts for Training \fullname}
\label{app:prompts_for_training}
\begin{systemprompt}[label={lst:search-method-prompt}]{Example Object Search Prompt}
\begin{lstlisting}[style=jsonstyle]
[
  {
    "role": "system",
    "content": [
      {
        "text": "You are a robot program that predicts actions. The video input from the egocentric camera shows the most recent actions the robot has executed. The images are selected frames of particular importance from all the actions the robot has executed so far. Based on these, output the current subtask the robot should execute and nothing else.\n\nReturn a JSON with:\n- current_subtask: the action that should be executed at the current timestep\n- keyframe_positions: list of frame positions (1-indexed) from the video input where actions change\n"
      }
    ]
  },
  {
    "role": "user",
    "content": [
      {
        "text": "Task: The robot's wrist and third-person camera feed is shown below. What subtask should the robot execute to retrieve the fried chicken and put it in the white bin?\nHere are the selected frames from the entirety of the full video that are of particular importance:"
      },
      {
        "image": "<path_to_memory_frame_1.png>"
      },
      {
        "image": "<path_to_memory_frame_2.png>"
      },
      {
        "text": "\nHere is a video of the most recent actions the robot has executed:"
      },
      {
        "video": [
          "<path_to_context_frame_1.png>",
          "<path_to_context_frame_2.png>",
          "<path_to_context_frame_3.png>",
          "<path_to_context_frame_4.png>",
          "<path_to_context_frame_5.png>",
          "<path_to_context_frame_6.png>",
          "<path_to_context_frame_7.png>",
          "<path_to_context_frame_8.png>"
        ]
      }
    ]
  },
  {
    "role": "assistant",
    "content": [
      {
        "text": "{\"current_subtask\": \"take the fried chicken from the right bin and place in the white bin\", \"keyframe_positions\": [7]}"
      }
    ]
  }
]
\end{lstlisting}
\end{systemprompt}

\begin{systemprompt}[label={lst:counting-method-prompt}]{Example Counting Prompt}
\begin{lstlisting}[style=jsonstyle]
[
  {
    "role": "system",
    "content": [
      {
        "text": "You are a robot program that predicts actions. The video input from the egocentric camera shows the most recent actions the robot has executed. The images are selected frames of particular importance from all the actions the robot has executed so far. Based on these, output the current subtask the robot should execute and nothing else.\n\nReturn a JSON with:\n- current_subtask: the action that should be executed at the current timestep\n- keyframe_positions: list of frame positions (1-indexed) from the video input where actions change\n"
      }
    ]
  },
  {
    "role": "user",
    "content": [
      {
        "text": "Task: What subtask should the robot execute to get three scoops of peanuts and put it in the green bowl, and two scoops of jelly beans and put it in the blue bowl?\nHere are the selected frames from the entirety of the full video that are of particular importance:"
      },
      {
        "image": "<path_to_memory_frame.png>"
      },
      {
        "text": "\nHere is a video of the most recent actions the robot has executed:"
      },
      {
        "video": [
          "<path_to_context_frame_1.png>",
          "<path_to_context_frame_2.png>",
          "<path_to_context_frame_3.png>",
          "<path_to_context_frame_4.png>",
          "<path_to_context_frame_5.png>",
          "<path_to_context_frame_6.png>",
          "<path_to_context_frame_7.png>",
          "<path_to_context_frame_8.png>"
        ]
      }
    ]
  },
  {
    "role": "assistant",
    "content": [
      {
        "text": "{\"current_subtask\": \"reset scooper position\", \"keyframe_positions\": []}"
      }
    ]
  }
]
\end{lstlisting}
\end{systemprompt}

\begin{systemprompt}[label={lst:dusting-method-prompt}]{Example Dusting Prompt}
\begin{lstlisting}[style=jsonstyle]
[
  {
    "role": "system",
    "content": [
      {
        "text": "You are a robot program that predicts actions. The video input from the egocentric camera shows the most recent actions the robot has executed. The images are selected frames of particular importance from all the actions the robot has executed so far. Based on these, output the current subtask the robot should execute and nothing else.\n\nReturn a JSON with:\n- current_subtask: the action that should be executed at the current timestep\n- keyframe_positions: list of frame positions (1-indexed) from the video input where actions change\n"
      }
    ]
  },
  {
    "role": "user",
    "content": [
      {
        "text": "Task: What subtask should the robot execute to remove the items from the shelves, dust the shelves, and place the items back on the shelves?\nHere are the selected frames from the entirety of the full video that are of particular importance:"
      },
      {
        "image": "<path_to_memory_frame_1.png>"
      },
      {
        "image": "<path_to_memory_frame_2.png>"
      },
      {
        "image": "<path_to_memory_frame_3.png>"
      },
      {
        "image": "<path_to_memory_frame_4.png>"
      },
      {
        "text": "\nHere is a video of the most recent actions the robot has executed:"
      },
      {
        "video": [
          "<path_to_context_frame_1.png>",
          "<path_to_context_frame_2.png>",
          "<path_to_context_frame_3.png>",
          "<path_to_context_frame_4.png>",
          "<path_to_context_frame_5.png>",
          "<path_to_context_frame_6.png>",
          "<path_to_context_frame_7.png>",
          "<path_to_context_frame_8.png>"
        ]
      }
    ]
  },
  {
    "role": "assistant",
    "content": [
      {
        "text": "{\"current_subtask\": \"remove the object on the top shelf\", \"keyframe_positions\": [5]}"
      }
    ]
  }
]
\end{lstlisting}
\end{systemprompt}

\section{Prompts for GPT-5 / Gemini Robotics–ER 1.5 Evaluation}
\label{sec:gpt_prompt} 

\begin{systemprompt}[label={lst:search-prompt}]{Object Search System Prompt}
You are an AI assistant controlling a single-arm robot to search for specific objects amongst 3 bins. When exploring the bins for objects, look in the order of left bin, center bin, then right bin. You will receive images from two cameras: one for a global view and one on the robot's wrist for a detailed view. You will be provided with recent images that show the most recent actions the robot has executed. You will also be provided with selected keyframe images which are frames of particular importance from all the actions the robot has executed so far. Based on these, choose an action from the provided list for the robot to execute to best achieve the user's task instruction. Provide the exact action from the list without any explanation.

\medskip
You will select your action from the following list:
\begin{itemize}[nosep, leftmargin=*]
    \item \texttt{look inside the <LOCATION> bin}
    \item \texttt{take the <OBJECT> from the <LOCATION> bin and place it in the white bin}
\end{itemize}

\medskip
\texttt{<LOCATION>} is one of "left", "center", or "right".

\texttt{<OBJECT>} is one of "green tape", "red block", "corn", "baguette", "blue block", "fried chicken", "milk carton", "ketchup", "eraser", "grapes", "strawberry", "tomato", "pear", "wooden block", or "olive oil".

\medskip
You will also return a list of values from 1-8 to index which of the frames from the most recent actions seem to be of particular importance for the robot to remember. For this task, recalling what objects are in each bin is critical, so you should return a list of indices, if any, from the most recent frames that provides a good view of a bin. 

\medskip
Return a JSON with:
\begin{itemize}[nosep, leftmargin=*]
    \item \texttt{current\_subtask}: the action that should be executed at the current timestep, selected from the above list using the stated \texttt{<OBJECT>} and \texttt{<LOCATION>} values
    \item \texttt{keyframe\_positions}: list of frame positions from 1-8, if any, from the recent frames to keep track of which objects are in each bin
\end{itemize}
\end{systemprompt}

\begin{systemprompt}[label={lst:scooping-prompt}]{Counting System Prompt}
You are an AI assistant guiding a single-arm robot to obtain a specific amount of scoops of two different ingredients. You will reset the scooper between each scoop, and drop the scooper when all scoops across both ingredients have been obtained. You will receive images from two cameras: one for a global view and one on the robot's wrist for a detailed view. You will be provided with recent images that show the most recent actions the robot has executed. You will also be provided with selected keyframe images which are frames of particular importance from all the actions the robot has executed so far. Based on these, choose an action from the provided list for the robot to execute to best achieve the user's task instruction. Provide the exact action from the list without any explanation.

\medskip
You will select your action from the following list:
\begin{itemize}[nosep, leftmargin=*]
    \item \texttt{pick up the scooper}
    \item \texttt{place a scoop of <OBJECT> in the <COLOR> bowl}
    \item \texttt{reset scooper position}
    \item \texttt{drop the scooper}
\end{itemize}

\medskip
\texttt{<OBJECT>} is one of "peanuts" or "jelly beans".

\texttt{<COLOR>} is one of "green" or "blue".

\medskip
You will also return a list of values from 1-8 to index which of the frames from the most recent actions seem to be of particular importance for the robot to remember. For this task, recalling how many scoops of each ingredient have been obtained is critical, so you should return a list of indices, if any, from the most recent frames that provides a good view of a completed scoop. 

\medskip
Return a JSON with:
\begin{itemize}[nosep, leftmargin=*]
    \item \texttt{current\_subtask}: the action that should be executed at the current timestep, selected from the above list using the stated \texttt{<OBJECT>} and \texttt{<COLOR>} values
    \item \texttt{keyframe\_positions}: list of frame positions from 1-8, if any, from the recent frames to keep track of scoops
\end{itemize}
\end{systemprompt}

\begin{systemprompt}[label={lst:dusting-prompt}]{Dusting System Prompt}
You are an AI assistant guiding a single-arm robot to take an object off each shelf (bottom shelf then top shelf), pick up the duster, dust the bottom shelf, reset the duster, dust the top shelf, put down the duster, and replace the objects back to their original places (bottom shelf then top shelf). You will receive images from two cameras: one for a global view and one on the robot's wrist for a detailed view. You will be provided with recent images that show the most recent actions the robot has executed. You will also be provided with selected keyframe images which are frames of particular importance from all the actions the robot has executed so far. Based on these, choose an action from the provided list for the robot to execute to best achieve the user's task instruction. Provide the exact action from the list without any explanation.

\medskip
You will select your action from the following list:
\begin{itemize}[nosep, leftmargin=*]
    \item \texttt{remove the object on the bottom shelf}
    \item \texttt{remove the object on the top shelf}
    \item \texttt{pick up duster}
    \item \texttt{dust bottom shelf}
    \item \texttt{reset duster}
    \item \texttt{dust top shelf}
    \item \texttt{put down duster}
    \item \texttt{place the <OBJECT> on the bottom shelf}
    \item \texttt{place the <OBJECT> on the top shelf}
\end{itemize}

\medskip
\texttt{<OBJECT>} is one of "panda plushie", "purple plushie", "zebra plushie", "elephant plushie", "lion plushie", "smily face ball", "hello kitty plushie", "baby shoe", "milk carton".

\medskip
You will also return a list of values from 1-8 to index which of the frames from the most recent actions seem to be of particular importance for the robot to remember. For this task, recalling where the items were originally placed on the shelves and which shelves have been dusted is critical, so you should return a list of indices, if any, from the most recent frames that provides a good indication of either. 

\medskip
Return a JSON with:
\begin{itemize}[nosep, leftmargin=*]
    \item \texttt{current\_subtask}: the action that should be executed at the current timestep, selected from the above list using the stated \texttt{<OBJECT>} values
    \item \texttt{keyframe\_positions}: list of frame positions from 1-8, if any, from the recent frames to keep track of where the objects were originally placed on the shelves and which shelves have been dusted
\end{itemize}
\end{systemprompt}
\end{document}